\documentclass[letterpaper, 10 pt, conference]{ieeeconf}  

\IEEEoverridecommandlockouts                              

\usepackage{graphicx}
\usepackage{pifont}
\usepackage{amsmath}
\usepackage{amsfonts}
\usepackage{caption}
\usepackage{subcaption}
\usepackage{tabularx}
\usepackage{booktabs} 
\let\labelindent\relax
\usepackage{enumitem,kantlipsum} 
\usepackage{hyperref}
\usepackage[symbol]{footmisc}

\newcommand{\cmark}{\ding{51}}%
\newcommand{\xmark}{\ding{55}}%
\usepackage{xcolor}
\newcommand{\remove}[1]{}
\DeclareMathOperator*{\argmax}{argmax}

\title{\LARGE \bf
Hiding Leader's Identity in Leader-Follower Navigation\\
through Multi-Agent Reinforcement Learning
\vspace{-3mm}
}
\author{Ankur Deka$^{1}$, Wenhao Luo$^{1}$, Huao Li$^{2}$, Michael Lewis$^{2}$ and Katia Sycara$^{1}$
\vspace{-6mm}
\thanks{This research has been supported by ONR N00014-19-C-1070, AFOSR/AFRL award FA9550-18-1-0251, FA9550-18-1-0097 and AFOSR FA9550-15-1-0442. A video summary is attached. Code:  \href{https://github.com/Ankur-Deka/Hiding-Leader-Identity}{\texttt{github.com/Ankur-Deka/Hiding-Leader-Identity}}.} 
\thanks{$^{1}$Robotics Institute, Carnegie Mellon University, 5000 Forbes Ave., Pittsburgh, USA.
        {\tt\small \{adeka, luo, katia\}@cs.cmu.edu}}%
\thanks{$^{2}$School of Computing and Information, University of Pittsburgh, 4200 Fifth Ave., Pittsburgh, USA
        {\small {\tt hul52@pitt.edu} and {\tt ml@sis.pitt.edu}}}%
}

\begin{document}
\maketitle
\thispagestyle{empty}
\pagestyle{empty}
\begin{abstract}
Leader-follower navigation is a popular class of multi-robot algorithms where a leader robot leads the follower robots in a team. The leader has specialized capabilities or mission critical information (e.g. goal location) that the followers lack, and this makes the leader crucial for the mission's success. However, this also makes the leader a vulnerability - an external adversary who wishes to sabotage the robot team's mission can simply harm the leader and the whole robot team's mission would be compromised. Since robot motion generated by traditional leader-follower navigation algorithms can reveal the identity of the leader, we propose a defense mechanism of hiding the leader's identity by ensuring the leader moves in a way that behaviorally camouflages it with the followers, making it difficult for an adversary to identify the leader. To achieve this, we combine Multi-Agent Reinforcement Learning, Graph Neural Networks and adversarial training. Our approach enables the multi-robot team to optimize the primary task performance with leader motion similar to follower motion, behaviorally camouflaging it with the followers. Our algorithm outperforms existing work that tries to hide the leader's identity in a multi-robot team by tuning traditional leader-follower control parameters with Classical Genetic Algorithms. We also evaluated human performance in inferring the leader's identity and found that humans had lower accuracy when the robot team used our proposed navigation algorithm.
\end{abstract}


\section{INTRODUCTION}
\label{section:introduction}

Multi-robot systems are useful for many scenarios such as drone-delivery \cite{choudhury2021efficient}, agriculture \cite{albani2017monitoring}, search-and-rescue \cite{queralta2020collaborative}, disaster relief \cite{gregory2016application} and defense \cite{deka2021natural}. A class of multi-robot algorithms called leader-follower navigation is particularly popular as it simplifies the task of controlling multiple robots. It involves at least one leader robot which the other (follower) robots follow. Using leader follower navigation, it is sufficient to command only the leader robot and the follower robots simply follow their leader.

\begin{figure}
    \centering
    \includegraphics[width=0.8\columnwidth]{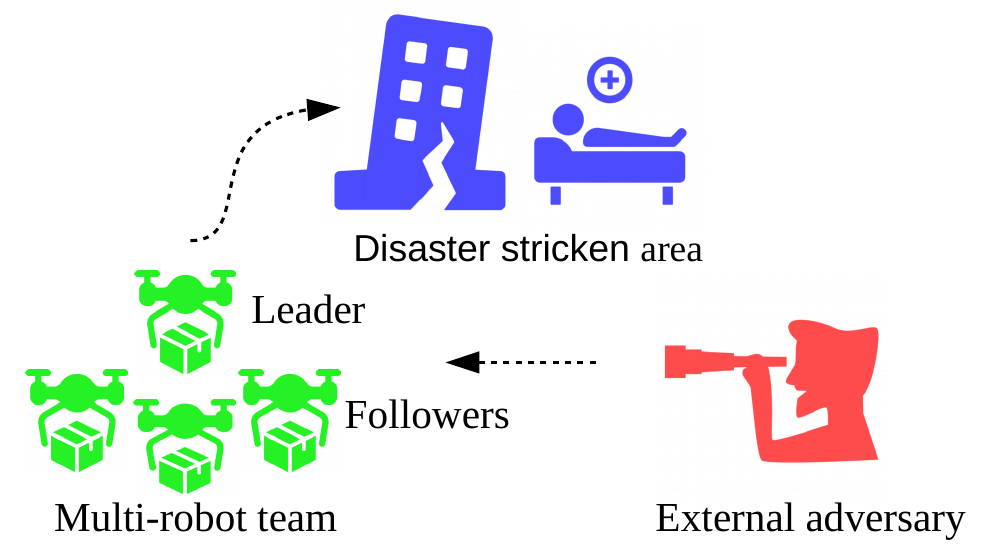}
    \caption{\small{A leader-follower multi-robot team in a disaster relief mission in the presence of an external adversary. It is crucial to hide the leader's identity from the adversary because if the adversary identifies and harms the leader then the whole team's mission would be compromised.}}
    \label{fig:disaster_relief}
\vspace{-8mm}
\end{figure}

Clearly, the leader is crucial for the robot team's success. Imagine a critical scenario such as a multi-robot team in a disaster relief mission as shown in Fig. \ref{fig:disaster_relief}. An external adversary (enemy) who wishes to sabotage the robot team's mission can simply identify and harm just the leader. This will compromise the whole robot team's mission. Thus, it is crucial to hide the leader's identity in such critical scenarios. Even if we make the visual appearance of the leader similar to the followers, it is possible to identify the leader by observing the motion of all the robots over time. E.g. the leader is usually ahead of the followers. An adversary can notice this and identify the leader.

We propose a defense mechanism of hiding the leader's identity by ensuring the leader moves in a way that behaviorally camouflages it with the followers, making it difficult for an adversary to identify the leader. Here, by ``behavioral camouflage" we refer to leader behavior (motion) that resembles followers' motion, blending it with the followers. Traditional leader-follower controllers are not good at hiding the leader's identity - including an approach \cite{zheng2020adversarial} that explicitly tried to hide the leader's identity, as we will see in Section \ref{section:results_swarm_performance}). Moreover, it is difficult to design a leader identity hiding multi-agent controller by hand, since such a controller would require complex multi-agent coordination.

Hence, instead of relying on traditional leader-follower controllers, we propose to leverage the recent advancements in Multi-Agent Reinforcement Learning (MARL), Graph Neural Networks (GNNs) and adversarial training. MARL allows us to define the objective of a multi-agent (in our case multi-robot) team through scalar reward signals given to the agents which the agents try to maximize. GNNs are deep learning architectures that allow us to model and train a large \& variable number of agents, which is appealing for real world applications. Adversarial training allows us to incorporate an intelligent artificial adversary that seeks to identify the leader.

We combine the power of MARL, GNNs and adversarial training for the task of hiding the leader's identity through a 3-stage training process. First, agents (robots) are trained with MARL to maximize a primary task reward, without caring about hiding the leader's identity. Then, robot trajectories (spatial coordinates over time) are collected from this multi-robot team. These trajectories are used as training data to train an artificial adversary with supervised learning to identify the leader. Finally, the agents are trained again with MARL - but this time they are given both a primary task reward and an identity hiding reward.  

We test our approach on a simulated multi-robot goal reaching task where the goal is known only to the leader and there is no communication between the robots. Using our proposed approach, the multi-robot team successfully reaches the goal while hiding the leader's identity from an artificial adversary, outperforming the baselines - both traditional leader-follower controller and MARL algorithms. We also evaluate human performance in inferring the leader's identity from multi-robot navigation videos and found that even humans found it hard to identify the leader when the multi-robot team deployed our proposed navigation algorithm.

The paper makes the following contributions: 
\begin{enumerate}
    \item We bring MARL, GNNs and adversarial training under one umbrella and present a multi-stage training process for the task of hiding the leader in a multi-robot team as a defense mechanism against an external adversary.
    \item We propose a novel deep learning architecture called Scalable-LSTM for modeling an artificial adversary.
    \item Our method is effective not only against an artificial adversary but it also "fools" human observers, i.e human observers do not detect the leader. To the best of our knowledge this is the first time this effect has been reported in the literature.
    \item We show that our approach can generalize to multi-robot teams with different sizes in a 0-shot fashion (wihout any fine tuning). 
\end{enumerate}

\section{Related Work}
Leader follower navigation is popular in the robotics community and has been studied in various contexts. \cite{sakai2017leader} addressed the issues of obstacle avoidance and connectivity preservation. \cite{vilca2016adaptive} proposed an adaptive strategy of formation reconfiguration. \cite{simonsenapplication} experimented with a leader-follower control algorithm on various mobile robots. All of these relied on traditional control based algorithms that didn't involve deep learning or Reinforcement Learning (RL).

Recent work has tried to incorporate RL into the leader-follower navigation problem \cite{miah2020model}. However, their method is designed for only 2-member robot teams with a single leader and a single follower. \cite{zhou2019adaptive} showed brief results with 2 followers and mentioned that they found it intractable to train a large number of agents due to exponentially growing state and action spaces as robot team size increases. 

We wish to not only leverage the power of RL but also have a large robot team with multiple followers. Because of this, we build up on our previous work on Multi Agent Reinforcement Learning (MARL) with Graph Neural Networks (GNNs) \cite{deka2021natural}. We have previously shown that by combining MARL \& GNNs large multi-agent teams can be tractably trained with reasonable computational resources (even a commodity laptop). In this work, we suitably modify the GNNs architecture to incorporate the constraints of the leader-follower navigation problem, as delineated in Section \ref{section:GNNs_multi_agent_architecture}. 

Once we have multiple robots in the leader-follower problem, protecting the leader's identity becomes crucial, as we have already discussed in Section \ref{section:introduction}. The importance of hiding the leader's identity has received limited attention in the past. We only found  \cite{zheng2020adversarial} to have attempted to address this issue. They used a traditional leader-follower controller and tuned its parameters with classical Genetic Algorithms (GAs) to hide leader's identity from an artificial adversary - a Convolutional Neural Network (CNN). However their results have limited evaluation. The adversary isn't benchmarked against other adversaries, which could mean that the adversary isn't smart enough and hence is easily deceived. In our experiments we saw that their approach couldn't hide the leader's identity when humans (instead of their artificial adversary) tried to identify the leader.

We instead propose a novel adversary architecture called Scalable LSTM (Long Short Term Memory) in Section \ref{section:adversary_architecture} and show its superiority over existing deep learning architectures in Section \ref{section:adversary_performance}. We experimentally show that our approach, relying on MARL \& GNNs rather than traditional leader-follower controllers, outperforms existing approach, \cite{zheng2020adversarial} that tried to hide the leader's identity. Further, we show that even humans have low accuracy in identifying the leader in a multi-robot team that navigates using our proposed approach.

\section{Problem Statement}
\begin{figure}
    \centering
    \vspace{2mm}
    \includegraphics[width=0.5\columnwidth]{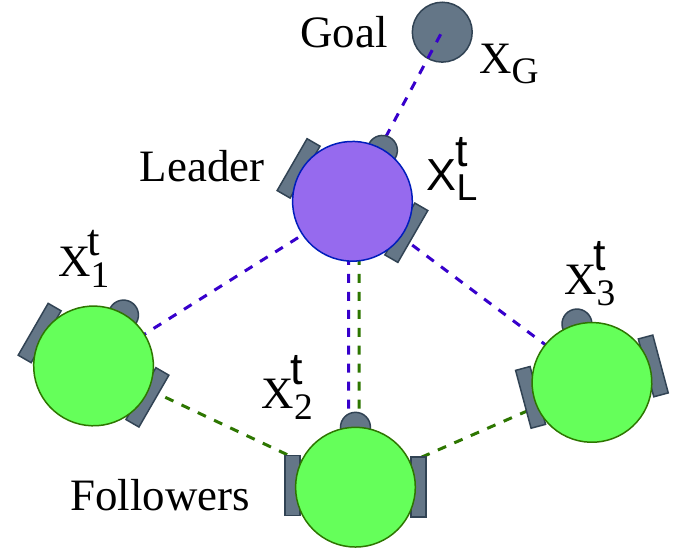}
    \caption{\small{A goal reaching task for a leader-follower multi-robot team. All robots can sense the neighboring robots and additionally the leader knows the goal location. All robots have the same visual appearance, the leader is shown in a different color only for illustrative purpose.}}
    \label{fig:problem_statement}
\vspace{-7mm}
\end{figure}

There is a n-robot team with $1$ leader, $L$ and $(n-1)$ followers, as shown in Fig. \ref{fig:problem_statement} with no communication between them. At every time step $t$, each robot can sense its own state, $X_i^t$  and the states of its neighbors, $\{X_{j}^t \mid j\neq i\}$ using on-board sensors. At the start of the mission, the goal is randomly located in the environment and only the leader knows its location $X_G$. E.g. in Fig. \ref{fig:problem_statement} the leader robot knows the goal and senses the states of all the neighboring followers; follower robot 2 only senses the states of the neighboring followers (to its left and right) and the leader.

Our objectives are (i) decentralized control for navigating the leader-follower multi-robot team to the goal location; and (ii) hide the leader's identity from an external adversary.
\begin{figure*}[ht]
    \vspace{2mm}
     \centering
     \begin{subfigure}[b]{0.25\textwidth}
         \centering
         \includegraphics[width=\textwidth]{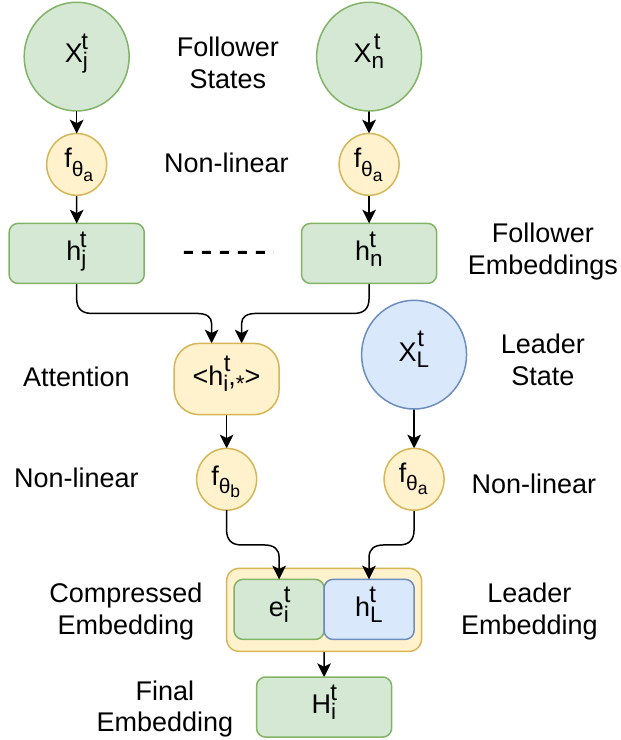}
         \caption{Follower architecture}
         \label{fig:follower_model}
     \end{subfigure}
     \quad
     \begin{subfigure}[b]{0.25\textwidth}
         \centering
         \includegraphics[width=\textwidth]{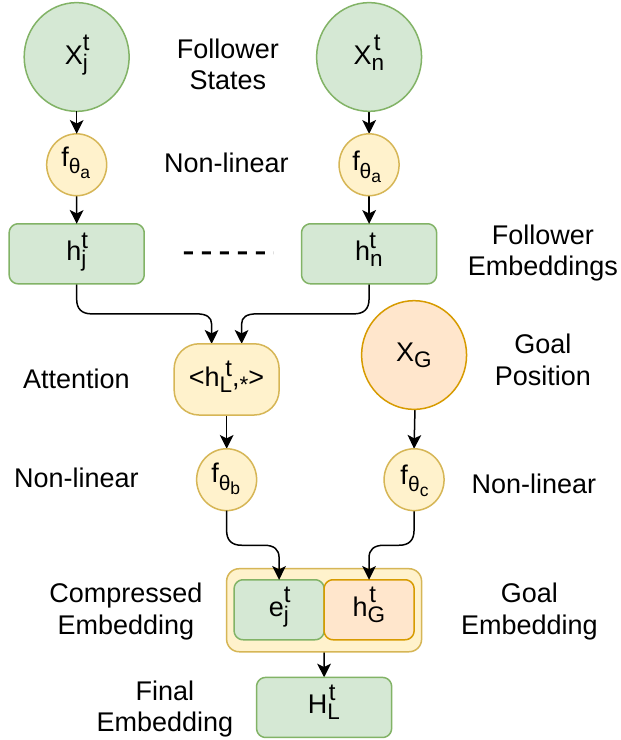}
         \caption{Leader architecture}
         \label{fig:leader_model}
     \end{subfigure}
     \quad
     \begin{subfigure}[b]{0.3\textwidth}
         \centering
         \includegraphics[width=0.83\textwidth]{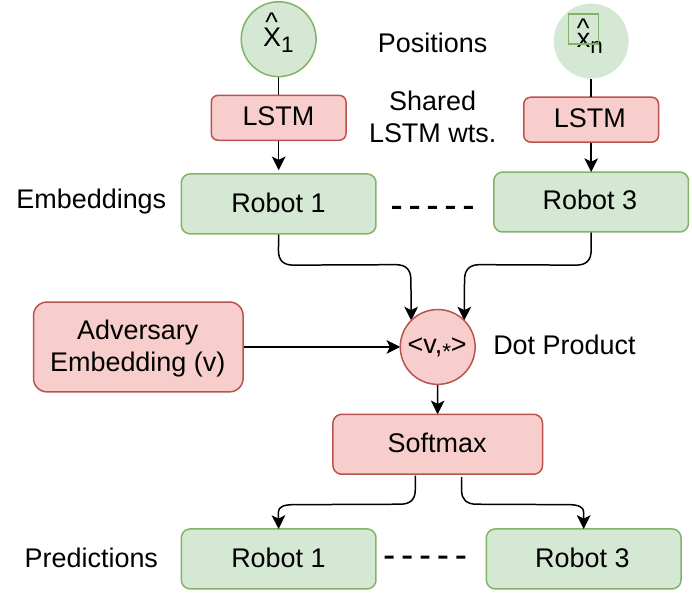}
         \caption{Scalable LSTM adversary architecture}
         \label{fig:adversary_model}
     \end{subfigure}
        \caption{\small GNNs based multi-agent architecture (Fig. \ref{fig:follower_model},\ref{fig:leader_model}) and Scalable-LSTM adversary architecture (Fig. \ref{fig:adversary_model}).
        }
        \label{fig:three graphs}
        \vspace{-5mm}
\end{figure*}

\section{Method}
We will use the terms \emph{robot} and \emph{agent} interchangeably. Let $Q=\{1,2,\dots,n\}$ denote the set of all agents and $S=Q-\{L\}$ denote the set of followers (excluding the leader). $\mathcal{A}$ denotes the external artificial adversary. We now describe the various components of our method. 
\subsection{MARL Formulation}
\label{section:marl_formulation}
The problem is formulated as a Decentralized Partially Observable Markov Decision
Process (Dec-POMDP), \cite{amato2013decentralized} with the following components:
\subsubsection{States and observations}
The state of $i^{th}$ agent at time step $t$ is denoted by $X_i^t \in \mathbb{R}^4$, which consists of its planar position and velocity. 
Each agent can observe it's own state $X_i^t$ and the states of its neighbors $\{X_{j}^t \mid j\neq i\}$. Additionally, the leader observes the goal location, $X_G$. 
$X^t=\{X_i^t \mid \forall i \in Q\}$ denotes the collective state of all agents at time $t$. $\bar{X}^t = X^t \bigcup X_G$ denotes the full environment state at time $t$. 
\subsubsection{Action space}
Each agent has an action space $a_i^t\in A_i$ that consists of accelerating in the $\pm x$ or $\pm y$ direction or not accelerating at all.
\subsubsection{Policy}
Each agent follows a policy $\pi_i$ to select its actions $a_i^t \sim \pi_i$. The leader's policy $a_L^t \sim \pi_L(a\mid \bar{X}^t)$ is conditioned on $\bar{X}^t = X^t \bigcup X_G$ which contains goal information. On the other other hand each follower's policy $a_i^t \sim \pi_i(a\mid X^t)$ is conditioned on $X^t$ which does not contain goal information. 
\subsubsection{Reward structure}
The primary task reward for agent $i$ at time $t$ is:
\begin{align}
    r_i^t &= \lambda_r \left(||X_G-\hat{X}_i^{t-1}||_2-||X_G-\hat{X}_i^{t}||_2\right) \label{eq:primary_task_reward}
\end{align}
Here $\hat{X}^t_i$ denotes the position of the agent and $\lambda_r=100$ is a scaling factor. This reward indicates how much closer the agent reached towards the goal in the current time step. Note that the same reward structure is used for both the leader and the followers. The reward encourages the leader to take large steps towards the goal. The followers don't have access to the goal location since their policy is not conditioned on $X_G$. They are indirectly encouraged to follow the leader which in turn will lead them to the goal.

The standard MARL objective (Naive MARL) for this reward signal is to maximize the cumulative reward, Eq. \ref{eq:naive_swarm_objective}.
Here, $\gamma=0.99$ is the discount factor. We add an additional identity hiding reward, $\mu_i^t$ which is indicative of how well the multi-robot team is hiding its leader's identity, Eq. \ref{eq:swarm_objective}.
\begin{align}
    \text{Objective}_\text{Naive MARL} &\equiv \max \mathbb{E}\left[ \sum_{i=1}^n\sum_{t\geq0} \gamma^t r_i^t\right]  
    \label{eq:naive_swarm_objective}\\
    \text{Objective}_\text{Proposed} &\equiv \max \mathbb{E}\left[ \sum_{i=1}^n\sum_{t\geq0} \gamma^t (r_i^t+\lambda_\mu \mu_i^t)\right]  
    \label{eq:swarm_objective}
\end{align}
$\lambda_\mu=1$ is a scaling factor controlling the importance of hiding leader's identity over the primary task. Computation of $\mu_i^t$ is described in the following sub-section in Eq. \ref{eq:privacy_reward}. Note that the rewards $r_i^t, \mu_i^t$ are only required for training and are not accessible to the agents at test time. 

\subsection{Adversarial Training}
\label{section:adversarial_learning}
The adversary's goal is to identify the ID of the leader from $i=\{1,2,\dots,n\}$ since there are n agents. the adversary ($\mathcal{A}$) observes the positions of all the agents $\hat{X}_i^t$ up to the current time step $O^t=\{\hat{X}_i^k| i \in Q, k\leq t\}$ and predicts the leader ID $l_{pred}^t$. 
\begin{align}
    l_{pred}^t &= \mathcal{A}(O^t) = \argmax_i P_\mathcal{A}(i\mid O^t)\\
    \mu_i^t &= -\mathbb{I}(l_{pred}^t=L)
    \label{eq:privacy_reward}
\end{align}
The adversary is trained with multi-class (since one amongst the $n$ agents is the leader) cross entropy loss which tries to match the predicted leader ID $l_{pred}^t$ to the true leader ID $L$. Once the adversary is trained, its prediction is used to generate identity hiding reward, $\mu_i^t$ (Eq. \ref{eq:privacy_reward}) as a feedback signal to the multi-robot team (Eq. \ref{eq:swarm_objective}). The negative sign in Eq. \ref{eq:privacy_reward} denotes that 
all agents get the same negative identity hiding reward, i.e. if the leader is identified by the adversary all the agents are penalized. This encourages cooperation and team spirit in the multi-agent team to jointly deceive the adversary by coordinating their motion.

\subsection{Graph Neural Networks Multi-agent Architecture}
\label{section:GNNs_multi_agent_architecture}
The agents (robots) in the multi-robot team can be treated as nodes of a graph to leverage the power of Graph Neural Networks (GNNs). GNNs are deep-learning architectures where the computations at the nodes and edges of the graph are performed by neural networks (parameterized non-linear functions), \cite{deka2018adaptive, deka2021natural}. Due to the presence of graph structure and multiple neural networks, they are called GNNs.

We incorporate the constraints of the leader-follower problem by ensuring that the goal information is accessible only to the leader and not the followers. In the following we describe the computations performed by an arbitrary follower and the leader.
\label{section:swarm_architecture}
\subsubsection{Followers}
Computation performed by an arbitrary follower agent $i$ is pictorially shown in Fig. \ref{fig:follower_model}. At every time step $t$, it takes as input its state, $X_i^t$ and passes it through a non-linear function, $f_{\theta_a}$ to compute an embedding, $h_i^t$. Similarly, it computes embeddings for all the other neighboring agents.
\begin{align}
    h_i^t &= f_{\theta_a}(X_i^t) \quad \forall i \in Q \label{eq:embed}
\end{align}

Follower agent $i$ then computes scaled dot-product attention \cite{vaswani2017attention}, $\psi_{ij}$ with all the other neighboring followers $v_j$'s.
\begin{align}
    \hat{\psi}_{ij}^t &= \frac{1}{\sqrt{d}}<h_i^t,h_j^t>  \quad \forall j \in S, j \neq i\\
    \psi_{ij}^t &= \frac{\exp(\hat{\psi}_{ij}^t)}{\sum_{k \in S,k \neq i} \exp(\hat{\psi}_{ik}^t)}\label{eq:attention_normalization}\\
    m_i^t &= \sum_{j \in S, j \neq i} \psi_{ij}^th_j^t\\
    e_i^t &= f_{\theta_b}(\text{concat}(h_i^t, m_i^t))
\end{align}
$d$ is the dimension of the vectors in dot product $<,>$. $\psi_{ij}^t$ denotes the attention paid by agent follower agent $i$ to follower agent $j$ at time $t$. The total attention paid to the neighbors sums to 1 due to the normalization in Eq. \ref{eq:attention_normalization}.
It then concatenates $e_i^t$ with leader embedding, $h_L^t$ to compute its final embedding, $H_1^t$.
\begin{align}
    H_i^t &= \text{concat}(e_i^t, h_L^t)
\end{align}
\subsubsection{Leader}
The leader's computation is pictorially shown in Fig. \ref{fig:leader_model}. It is similar to followers' computations with the key differnces being (i) Leader state ($X_L^t$) is replaced by goal position ($X_G$) and (ii) $f_{\theta_c}$ is used instead of $f_{\theta_a}$ for computing goal embedding.

\subsubsection{Policy and Value function}
Once each follower computes its final embedding, it conditions its policy and value function on it.
\begin{align}
    \pi_i(a\mid X^t) &= f_{\theta_d}(a\mid H_i^t) \quad \forall i\in S \label{eq:follower_policy}\\
    V_{\pi_i}(X^t) &= f_{\theta_e}(H_i^t) \quad  \forall i\in S
\end{align}
The leader computes its policy and value function in a similar way but uses a different set of parameters.
\begin{align}
    \pi_L(a\mid \bar{X}^t) &= f_{\theta_f}(a\mid H_L^t)\\
    V_{\pi_L}(\bar{X}^t) &= f_{\theta_g}(H_L^t)
\end{align}

\subsection{Scalable LSTM Adversary Architecture}
\label{section:adversary_architecture}
Long Short Term Memory (LSTM), \cite{hochreiter1997long} is a memory based deep learning architecture that can work with sequential data. It takes data at current time step as input and retains the past data in memory. It is temporally adaptable since it can handle input of any time duration.

However, it is not adaptable to different number of agents. This is because, the input \& output sizes at time $t$ depend on the number of agents $n$ and an LSTM can be designed to handle only fixed sizes of input \& output. If we directly feed the positions of all agents at time $t$ to an LSTM, the input size would be $2\times n$ (planar position) and the output size would be $n$ (probability of each agent being the leader). 

We propose Scalable-LSTM adversary architecture which builds on top of LSTM. It inputs the position of each agent separately (to handle variable size inputs) and has a dot product operation (to handle variable size outputs). 

At every time step $t$, the adversary takes as input the position of each agent separately, $\hat{X}_i^t$ and passes it through an LSTM to compute an embedding $\hat{h}_i^t$. Therefore, for $n$ agents there would be $n$ passes through the same LSTM as follows:
\begin{align}
    \hat{h}_i^t, \hat{c}_i^t &= \text{LSTM}(\hat{X}_i^t,\hat{h}_i^{t-1},\hat{c}_i^{t-1}) \quad \forall i \in Q
\end{align}
Because the adversary does $n$ separate passes, the LSTM gets fixed size (=2) input irrespective of how large $n$ is.
Here, $\hat{h}_i^t$ and $\hat{c}_i^t$ denote the hidden state and cell state of LSTM at time $t$.

The adversary then computes pairwise dot product between each agent's embedding, $\hat{h}_i^t$ and an adversary embedding, $v$ (a trainable parameter vector) as in Eq. \ref{eq:adversary_dot_product}. The output is passed through soft-max function to compute its probabilistic belief of the leader's identity as in Eq. \ref{eq:adversary_belief}.
\begin{align}
    y_i^t &= <v,\hat{h}_i^t>  \quad \forall i \in Q \label{eq:adversary_dot_product}\\
    P_\mathcal{A}(i|O^t)  &= \frac{\exp(y_i^t)}{\sum_{m\in Q}\exp(y_m^t)} \quad \forall i \in Q\label{eq:adversary_belief} 
\end{align}
If we have n agents, there would be n dot-products $y_i^t=<v,\hat{h}_i^t>$ for $i=1$ to $n$ and correspondingly $n$ output probability values $P_\mathcal{A}(i|O^t)$, irrespective of dimension of $h^t_i$. Thus, number of outputs automatically scales with no. of agents. Here, $O^t$ denotes the observations of the adversary up to time step $t$, i.e. $O^t=\{\hat{X}_i^k| i \in Q, k\leq t\}$. 

The adversary architecture consists of two sets of trainable parameters, LSTM weights and a adversary embedding $v$ which we collectively denote as $\theta_{\mathcal{A}}$.

\begin{figure*}[ht]
     \vspace{1mm}
     \centering
     \begin{subfigure}[b]{0.3\textwidth}
         \centering
         \includegraphics[width=0.75\textwidth]{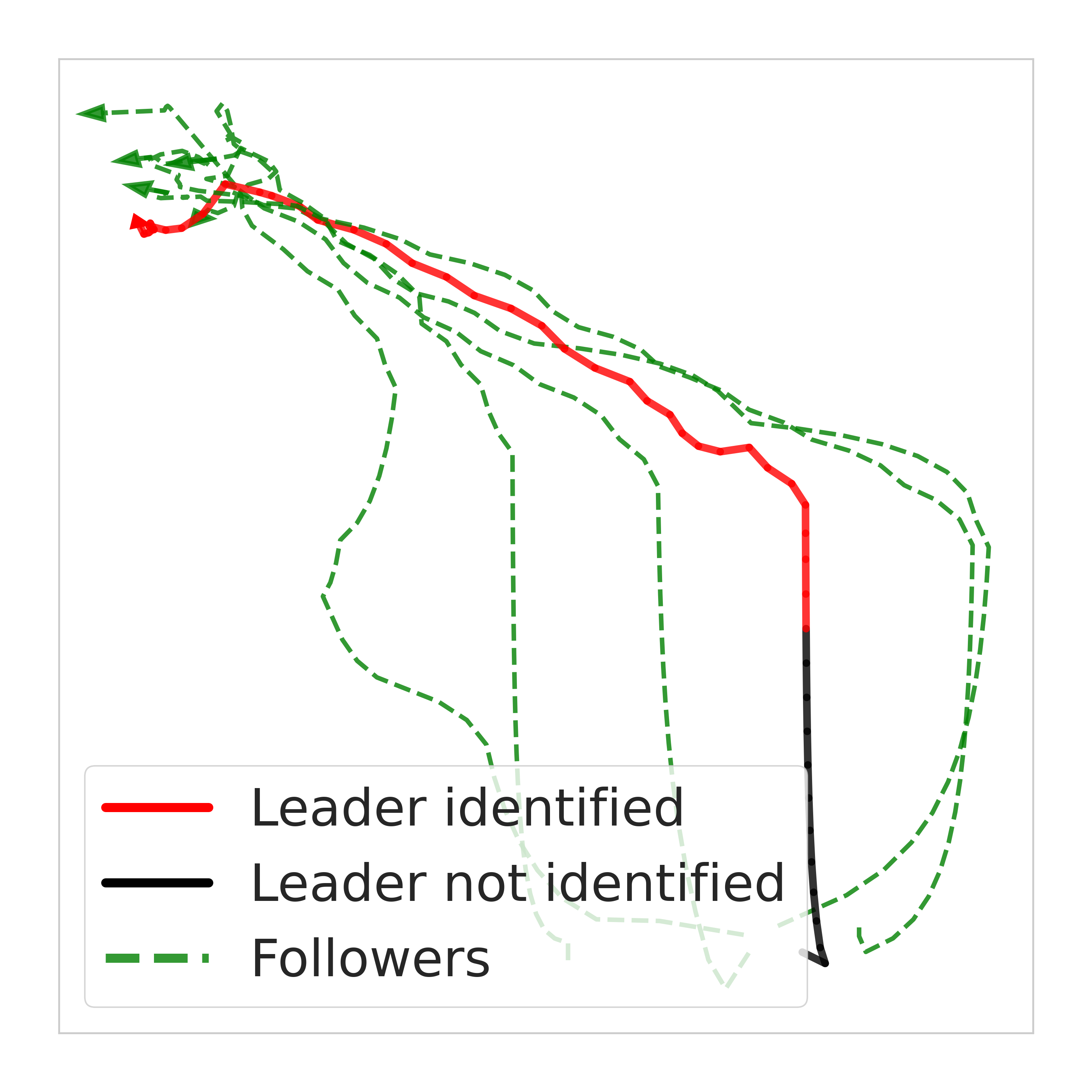}
         \vspace{-2mm}
         \caption{Naive MARL Multi-robot trajectories}
         \label{fig:multi_robot_trajectories}
     \end{subfigure}
     \quad \quad
     \begin{subfigure}[b]{0.7\columnwidth}
         \centering
         \includegraphics[width=\textwidth]{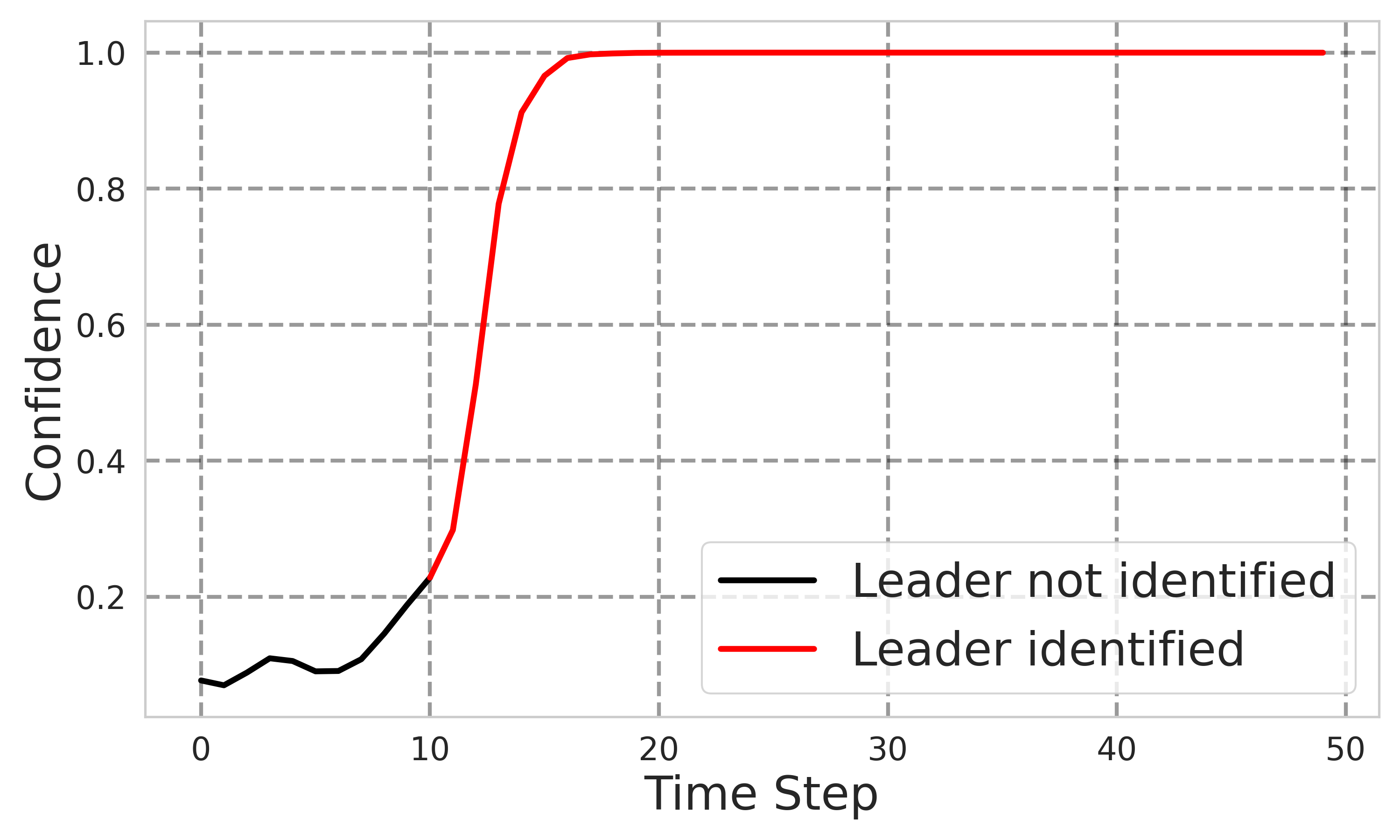}
         \vspace{-4mm}
         \caption{Adversary's confidence increases over time}
         \label{fig:adversary_confidence}
     \end{subfigure}
    \caption{\small An episode of Naive MARL Multi robot trajectories (Fig. \ref{fig:multi_robot_trajectories}) which the Scalable-LSTM adversary observes and tries to identify the leader. Scalable-LSTM predicts the leader at every time step $t$ based on the trajectory observation till time $t$. Initially it fails to identify the leader correctly (shown in black) but within 10 time steps in predicts leader correctly (shown in red). Fig. \ref{fig:adversary_confidence} shows confidence of the adversary on its prediction for the same episode. Again we can observe that initially Scalable LSTM has low confidence and fails to identify the leader (shown in black) but within 10 time it identifies the leader correctly (shown in red) and within 20 time steps its confidence is almost 1 ($100\%$ confident).
    }
    \label{fig:adversary_temporal_adaptability}
    \vspace{-6mm}
\end{figure*}

\subsection{Scalability, Adaptability and Decentralized Control}
\label{section:scalability}
The trainable parameters in our GNNs based Multi-agent architecture are $\theta = \{\theta_a,\theta_b,\dots,\theta_g\}$ of the non-linear functions $f_{\theta_a},f_{\theta_b},\dots,f_{\theta_g}$ respectively. We model these functions as neural networks, hence the name GNNs. Since the number of parameters is independent of the number of agents, the architecture is scalable and adaptable to different number of agents. The scalability and adaptability of Scalable LSTM adversary architecture is already described in the previous section (\ref{section:adversary_architecture}).  We validate these in our experimental results (Section \ref{section:adversary_performance} and \ref{section:swarm_generalization_num_agents}).

The parameters $\theta_a, \theta_b$ are shared across all agents and the paramters $\theta_d, \theta_e$ are shared across all the followers. This means that each agent maintains a separate copy of their respective parameters at test time for decentralized control.



\subsection{Training}
\label{section:training}
We train a leader-hiding multi-agent strategy in 3 stages:
\subsubsection{Training without identity hiding reward}
\label{section:training_state_1}
We first train a multi-agent team with the Naive MARL objective (Eq. \ref{eq:naive_swarm_objective}, doesn't include identity hiding reward $\mu_i^t$) using Proximal Policy Optimization (PPO) \cite{schulman2017proximal}, a popular model-free RL algorithm. The multi-agent team learns to navigate to the goal location without caring about hiding the leader's identity.

\subsubsection{Training an adversary}
\label{section:training_state_2}
Once we have a trained Naive MARL policy, we run the policy for multiple episodes and save the multi-robot trajectories along with the true leader ID in each episode. These  trajectories and leader ID's are provided to the adversary as training data. The adversary is trained with Stochastic Gradient Descent (SGD) supervised learning algorithm to predict the leader ID by minimizing the multi-class cross-entropy loss.
\subsubsection{Training with identity hiding reward}
\label{section:training_state_3}
Now that we have a trained adversary, we train a new multi-agent policy which tries to maximize both goal reaching reward and identity hiding reward (Eq. \ref{eq:swarm_objective} using PPO).

Due to the use of shared parameters in GNNs, our proposed approach is very simple to train with reasonable computational resources. We performed all our training on a commodity laptop with i7-7700HQ CPU, 16GB of RAM and a GTX 1060 GPU.\\
\textbf{Remark}\\
We would like to highlight that our formulation is consistent with the leader-follower problem because:
\begin{enumerate}[wide, labelwidth=!, labelindent=0pt]
    \item Followers are not given goal information as input. As shown in Eq. \ref{eq:follower_policy}, follower policies are conditioned on the embedding $H_i^t$ which doesn't contain any goal information. Goal information $X_G$ is completely absent from the GNNs based follower architecture, Fig. \ref{fig:follower_model}.
    \item Reward is NOT accessible at test time. Primary task reward is calculated using goal information, Eq. \ref{eq:primary_task_reward}. This might raise the confusion that the followers have access to goal information at test time. This is not the case as the agents have access to the reward only at training time and not at test time.
    \item Goal location is randomly changed in every episode. At the start of every episode, the goal is placed at a random location. The test goal location is different from the training location and it is impossible for the multi-agent policy to memorize a fixed goal-location seen during training.
\end{enumerate}
\vspace{-3mm}
\subsection{Environment}
The environment is a simulated goal reaching task built using Multi Agent Particle Environment (MAPE) \cite{lowe2017multi}.
\vspace{-3mm}
\section{Results}
\label{section:results}
\subsection{Adversary performance}
\label{section:adversary_performance}

\begin{table}[h!]
  \vspace{-3mm}
  \caption{Comparison of Adversary Architectures}
  \label{tab:adversary_comparision}
  \begin{center}
  \vspace{-3mm}
  \begin{tabular}{r|c|c|cc}
    \toprule
    Architecture & Accuracy & No. of & \multicolumn{2}{c}{Adaptability}\\
    & (max. 1) & Params. & Temporal & No. of agents\\
    \hline
    Random guess & 0.20 & 0 & \cmark & \cmark \\
    \hline
    LSTM & 0.95 & 2574 & \cmark & \xmark \\
    Zheng et al. \cite{zheng2020adversarial} & 0.97 &  109286 & \xmark & \xmark \\
    Scalable-LSTM & \textbf{0.99} & \textbf{936} & \cmark & \cmark \\
    \bottomrule
  \end{tabular}
  \vspace{-8mm}
  \end{center}
\end{table}
\begin{figure}[h!]
    \centering
    \includegraphics[width=0.9\columnwidth]{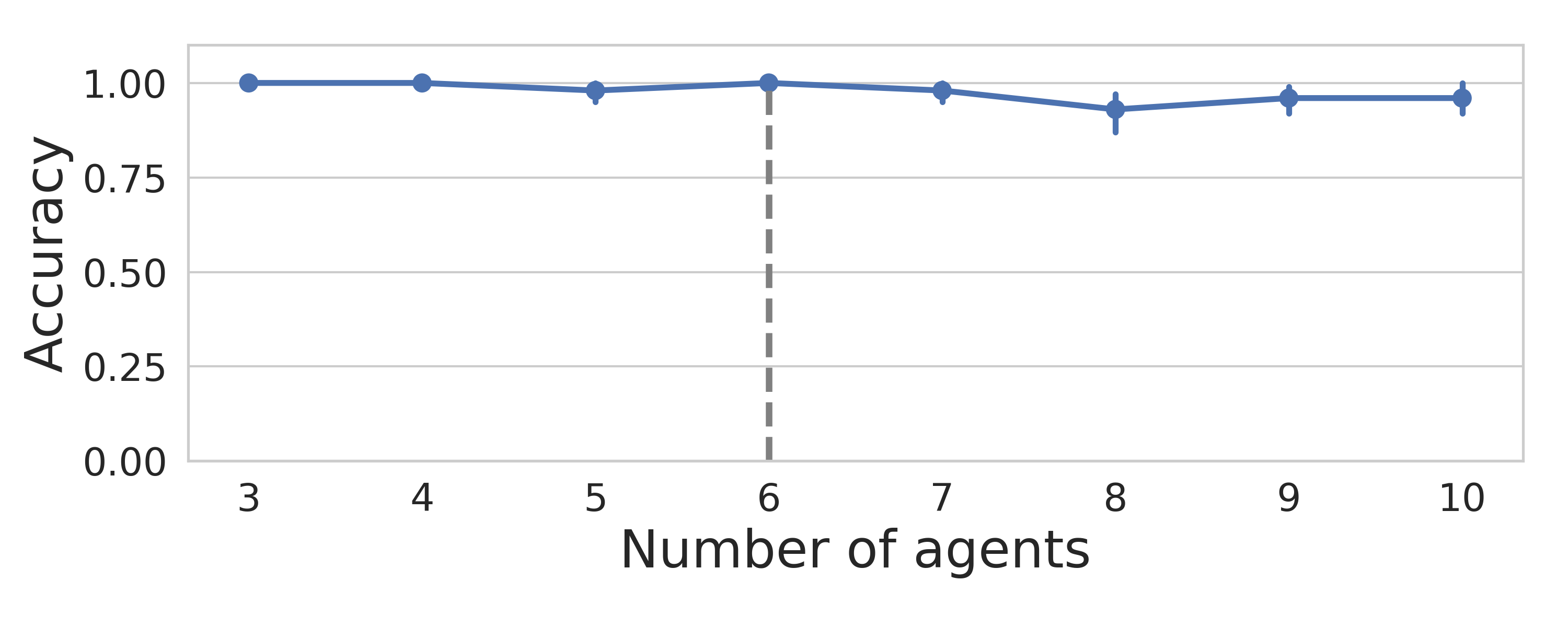}
    \vspace{-3mm}
    \caption{\small Scalable-LSTM adversary's 0-shot generalization to different number of agents. The blue curve, which shows the accuracy of Scalable-LSTM in identifying the leader in robot teams with different number of agents, constantly stays high (close to 1). Although Scalable-LSTM adversary was trained with only 6 agents (shown with dashed vertical line), it had high accuracy in identifying the leader in robot teams with number of agents varying between 3 and 10. }
    \label{fig:adversary_num_agents_generalization}
\end{figure}
\begin{figure*}[ht]
     \centering
     \begin{subfigure}[b]{0.35\textwidth}
         \centering
         \includegraphics[width=\textwidth]{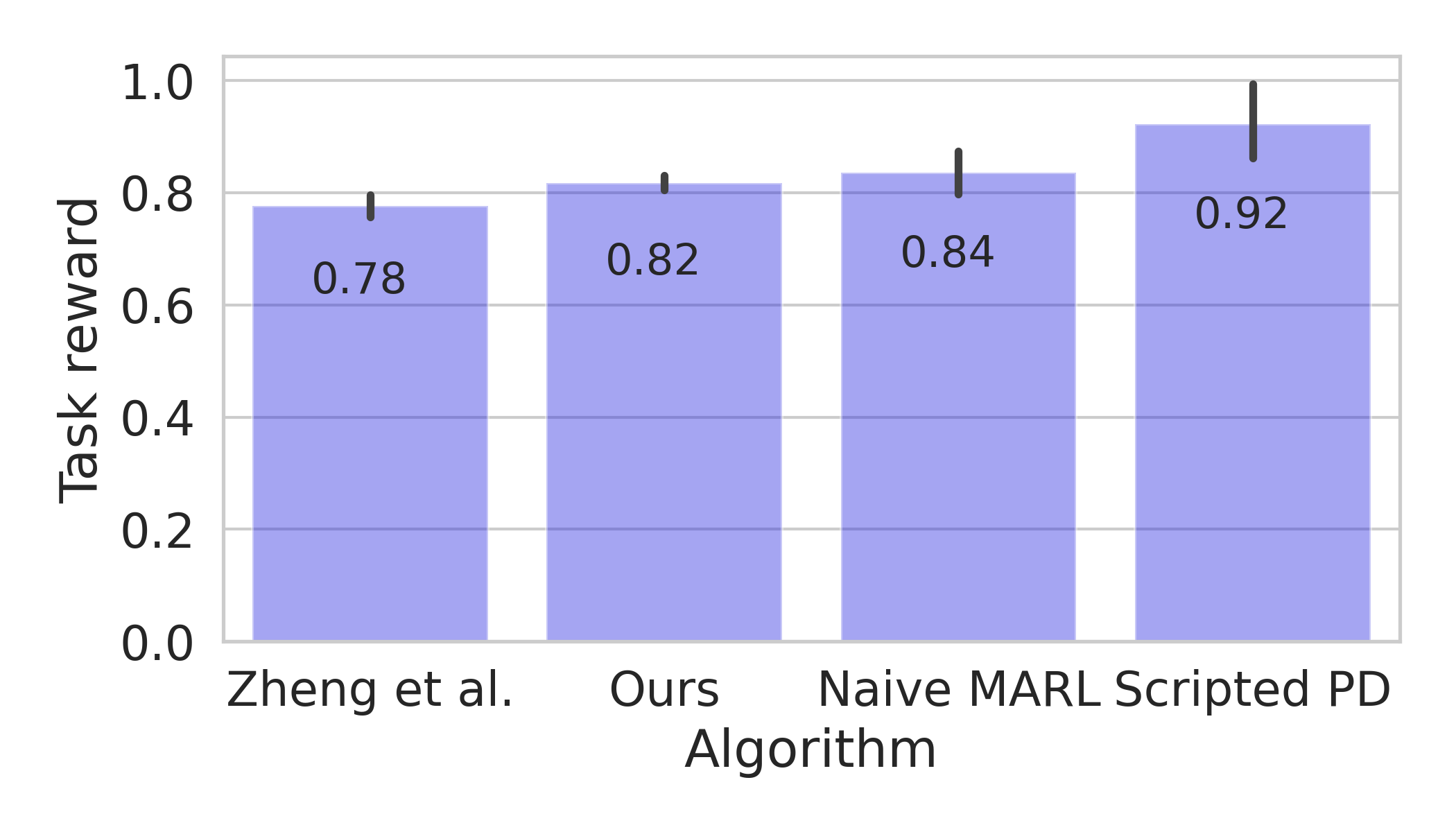}
         \vspace{-8mm}
         \caption{Primary task reward}
         \label{fig:task_reward}
     \end{subfigure}
     \begin{subfigure}[b]{0.35\textwidth}
         \centering
         \includegraphics[width=\textwidth]{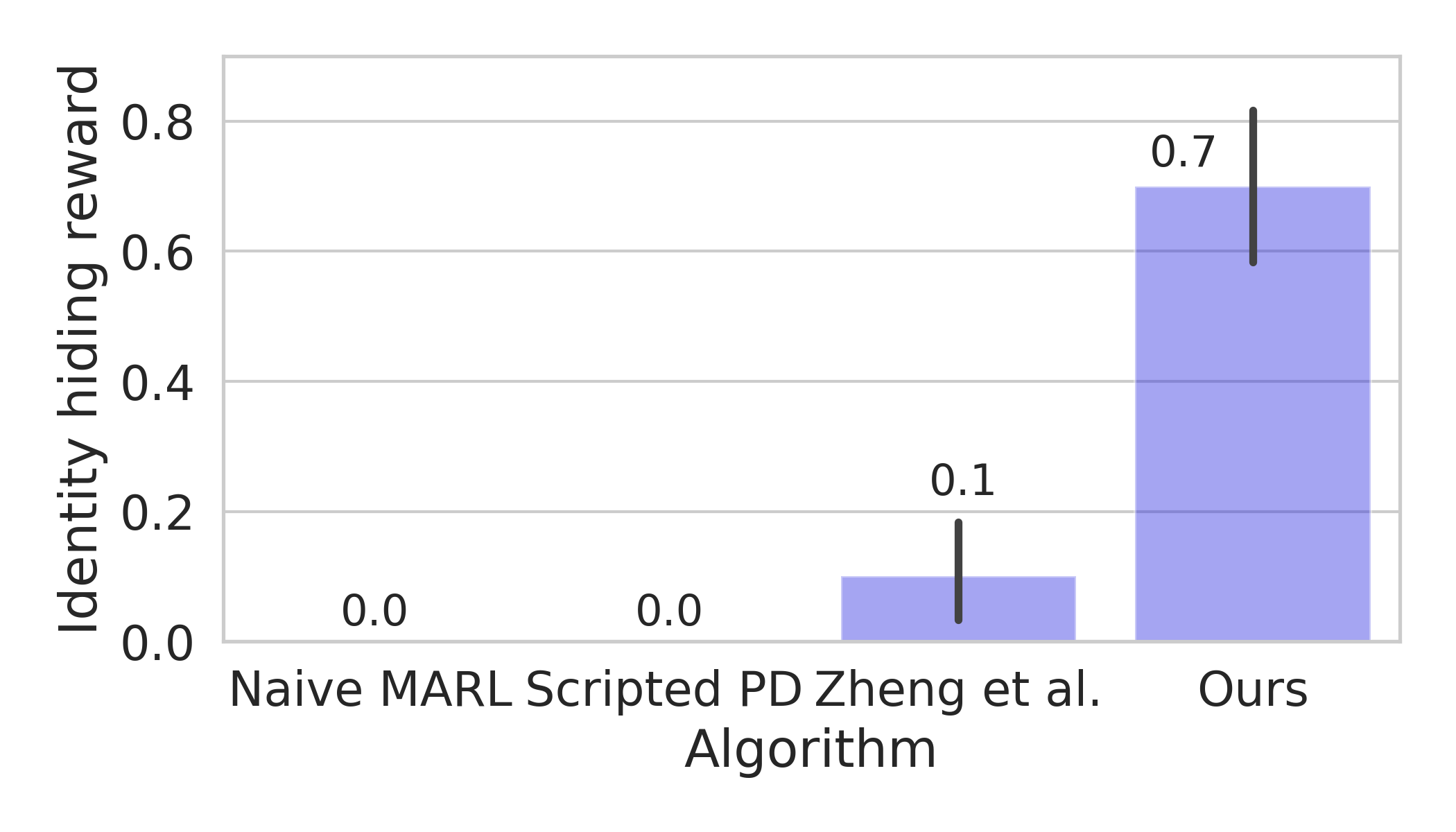}
         \vspace{-8mm}
         \caption{Identity hiding reward}
         \label{fig:privacy_reward}
     \end{subfigure}
    \vspace{-2mm}
    \caption{\small Multi-agent performance on primary task reward and identity hiding reward using different algorithms. Values are normalized between 0 and 1.}
    \label{fig:swarm_performance}
\end{figure*}

\newcommand{\w}{0.21\textwidth}
\begin{figure*}[ht]
    \setlength{\fboxsep}{0pt}%
     \centering
     \begin{subfigure}[b]{\textwidth}
         \centering
         \fbox{\includegraphics[width=\w]{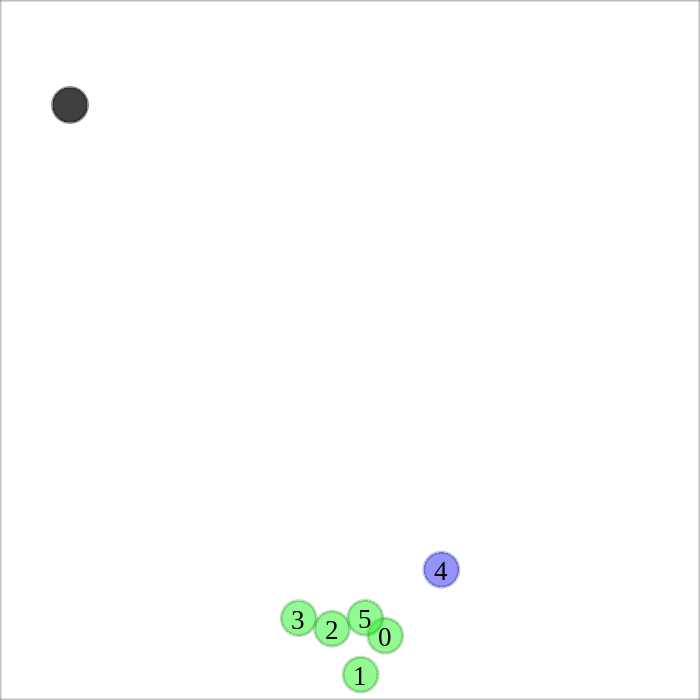}}
         \fbox{\includegraphics[width=\w]{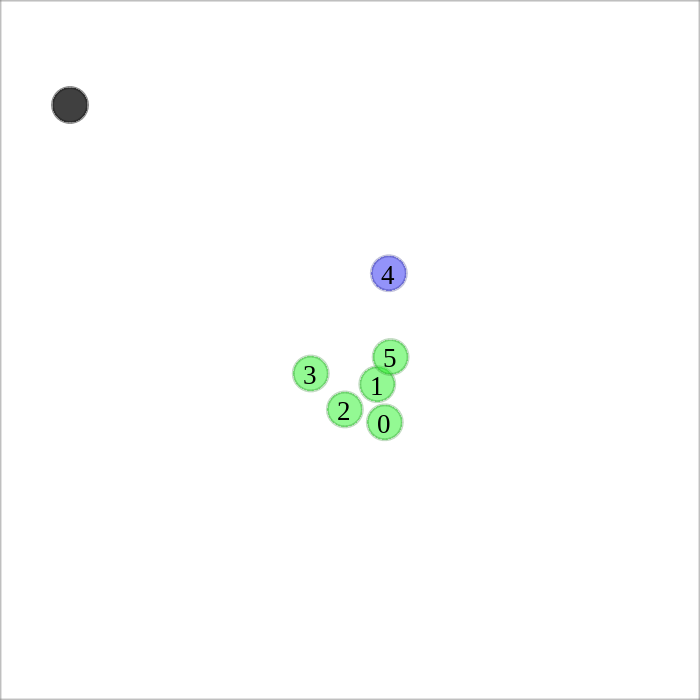}}
         \fbox{\includegraphics[width=\w]{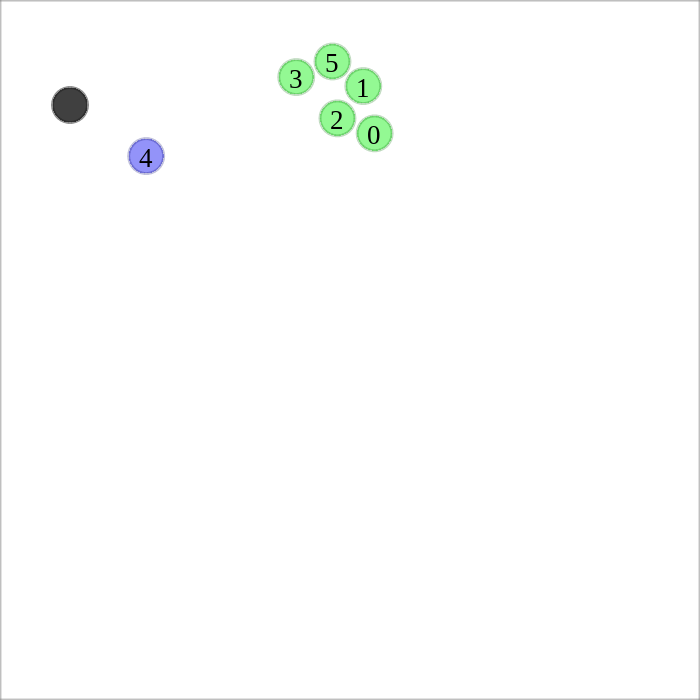}}
         \fbox{\includegraphics[width=\w]{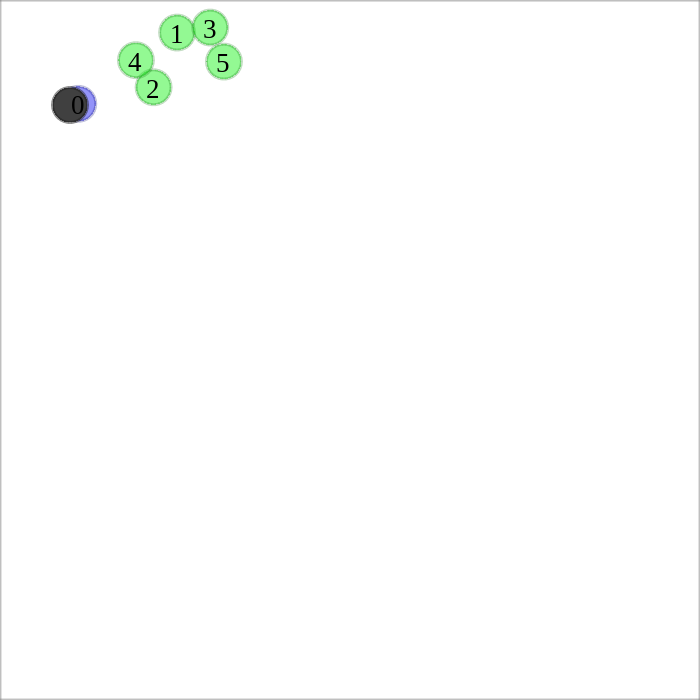}}
         \caption{Zheng et al. \cite{zheng2020adversarial}}
         \label{fig:zheng_trajectories}
     \end{subfigure}
     \begin{subfigure}[b]{\textwidth}
         \centering
         \fbox{\includegraphics[width=\w]{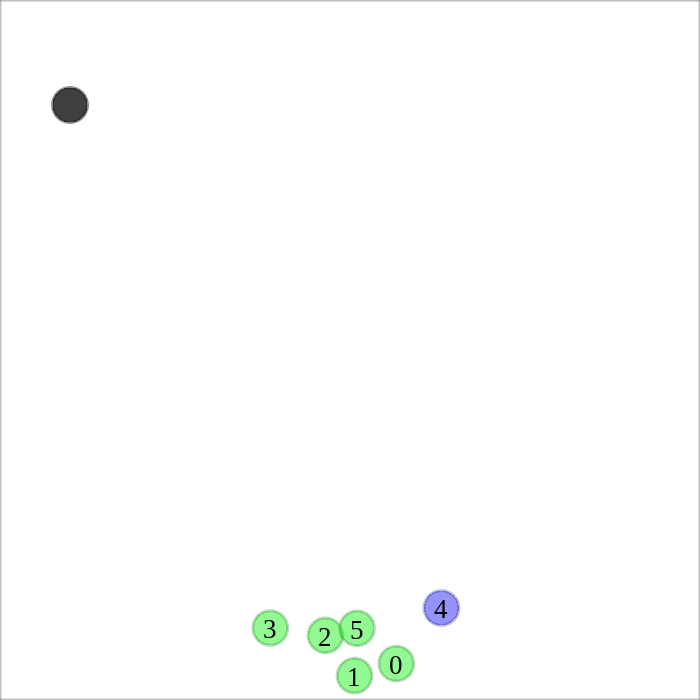}}
         \fbox{\includegraphics[width=\w]{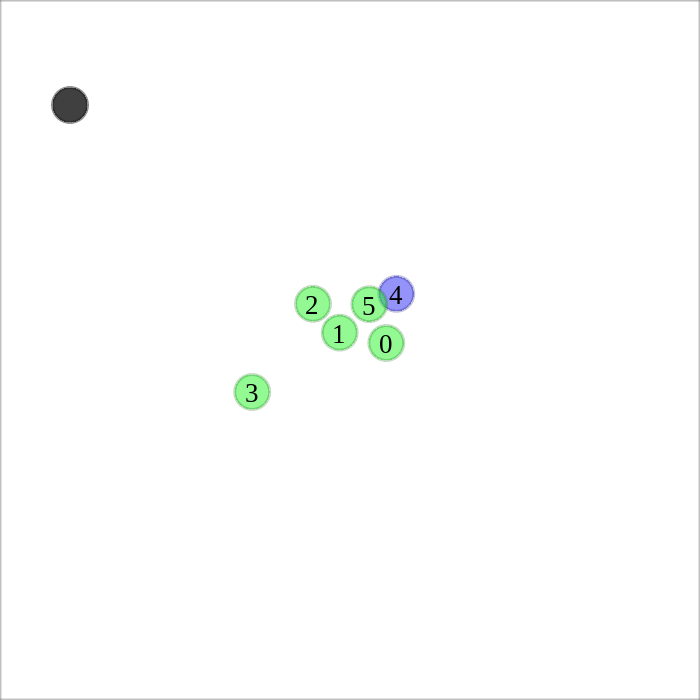}} 
         \fbox{\includegraphics[width=\w]{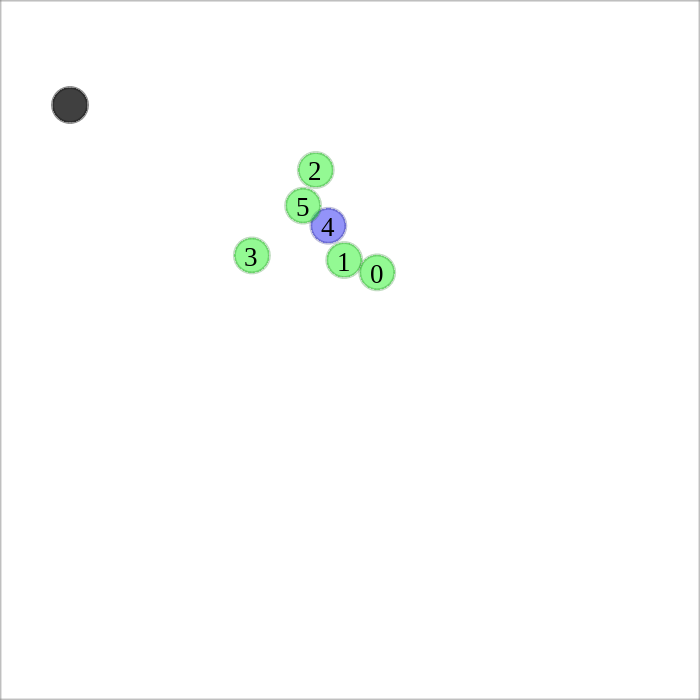}}
         \fbox{\includegraphics[width=\w]{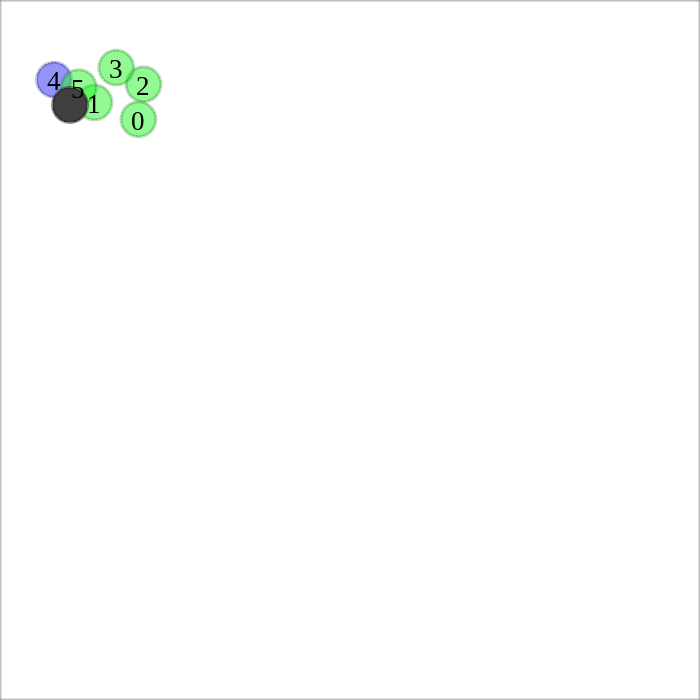}}
         \caption{Ours}
         \label{fig:our_trajectories}
     \end{subfigure}
     \vspace{-6mm}
    \caption{\small Multi-agent navigation using approach of Zheng et al. \cite{zheng2020adversarial} and our approach for the same goal location (shown as black circle). The leader is shown as a blue circle while the followers are shown as green circles. All agents have a number written on them - e.g. the leader (blue circle) is numbered 4. The leader's identity isn't concealed well by Zheng et al. (Fig. \ref{fig:zheng_trajectories}) as it is clearly ahead of the followers. Using our approach, the leader smartly moves with the followers as a group, hiding its identity (Fig. \ref{fig:our_trajectories}) - E.g. in 3rd snapshot of Fig. \ref{fig:our_trajectories}, a follower agent (numbered 3) deceptively seems to be leading the multi-agent team while the leader is behind. 
    }
    \label{fig:swarm_trajectories}
\end{figure*}
\newcommand{\ww}{0.27\textwidth}
\begin{figure*}[ht]
    \vspace{-3mm}
    \setlength{\fboxsep}{0pt}
     \centering
     \begin{subfigure}[b]{\ww}
         \centering
         \includegraphics[width=\textwidth]{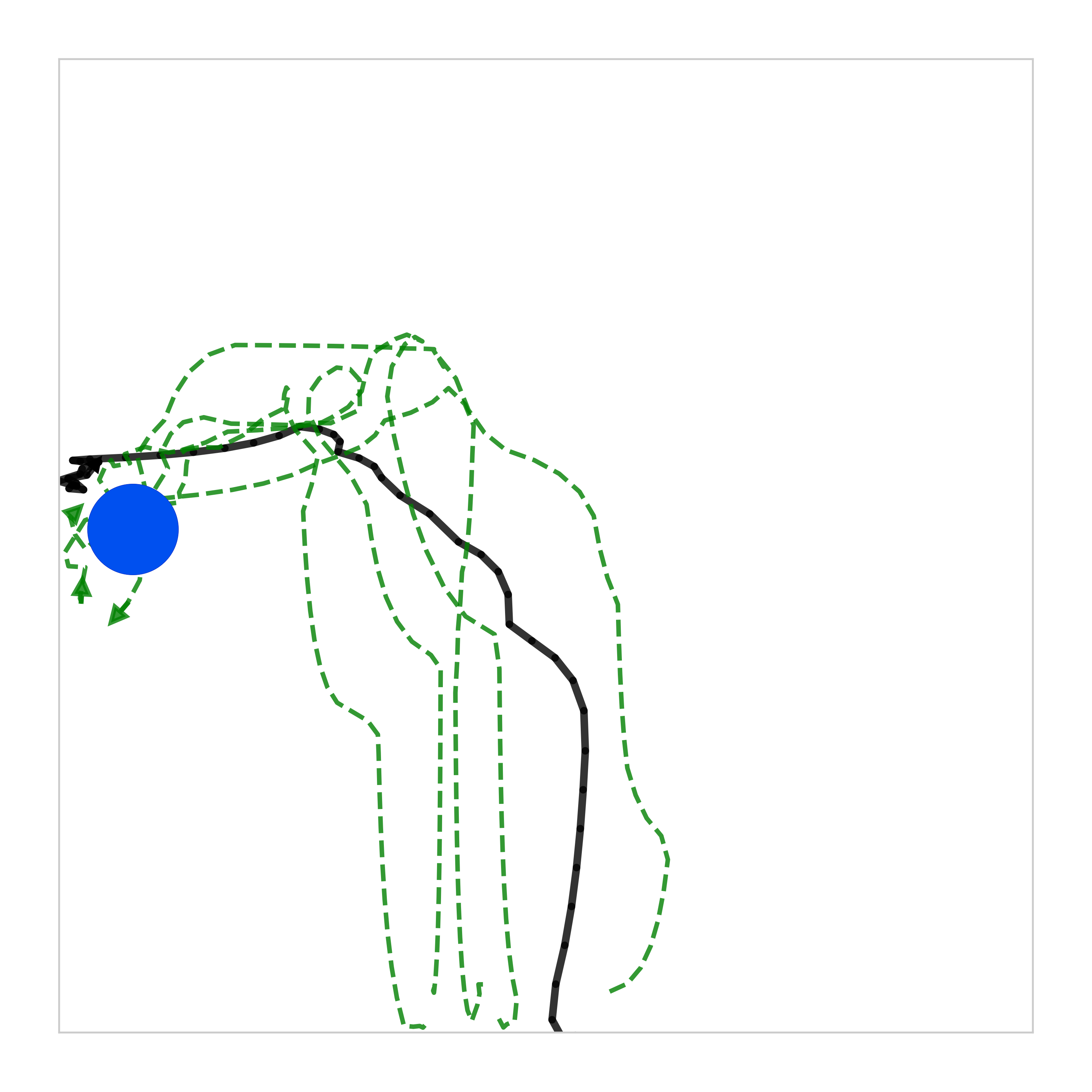}
         \vspace{-5mm}
         \caption{Goal location 1}
         \label{fig:score}
     \end{subfigure}
     \begin{subfigure}[b]{\ww}
         \centering
         \includegraphics[width=\textwidth]{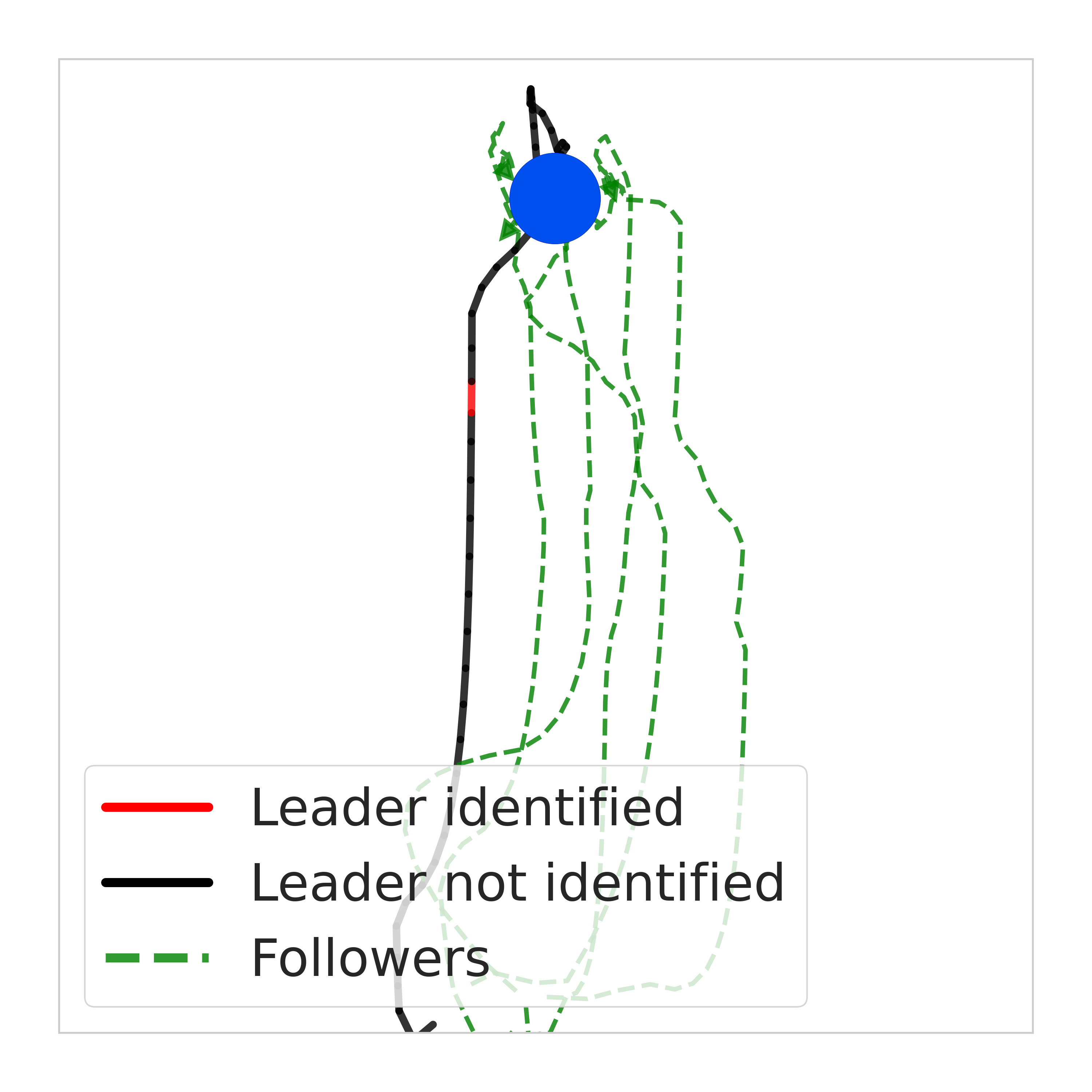}
         \vspace{-5mm}
         \caption{Goal Location 2}
         \label{fig:score}
     \end{subfigure}
     \begin{subfigure}[b]{\ww}
         \centering
         \includegraphics[width=\textwidth]{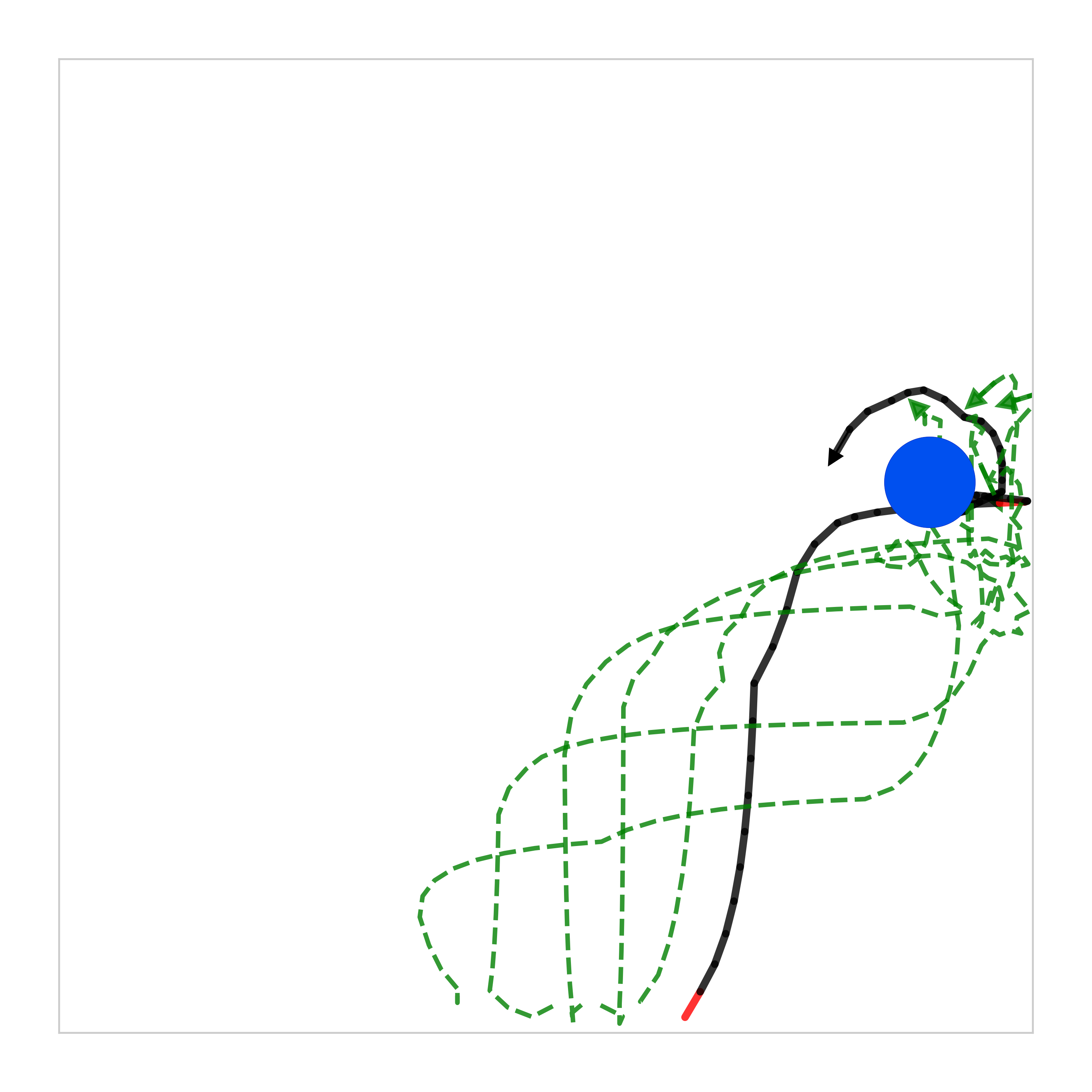}
         \vspace{-5mm}
         \caption{Goal location 3}
         \label{fig:score}
     \end{subfigure}
    \caption{Leader-follower navigation to different goal locations (shown as blue circle) using our proposed approach. The multi-agent team could successfully reach different goal locations while fooling the adversary which made a wrong prediction of the leader's identity as depicted by the black color of the leader trajectory in most regions.}
    \label{fig:different_goals}
\end{figure*}
\newcommand{\www}{0.2\textwidth}
\begin{figure*}[ht!]
    \vspace{3mm}
    \setlength{\fboxsep}{0pt}%
     \centering
         \fbox{\includegraphics[width=\www]{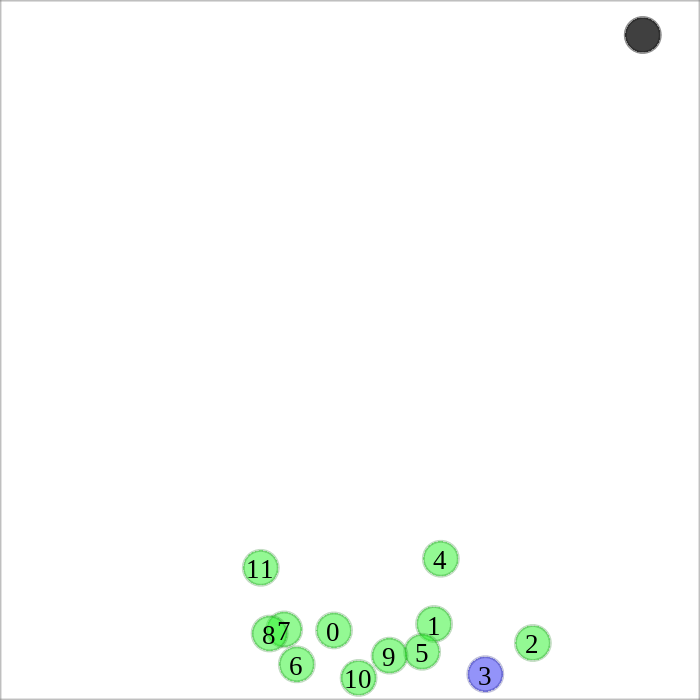}}
         \fbox{\includegraphics[width=\www]{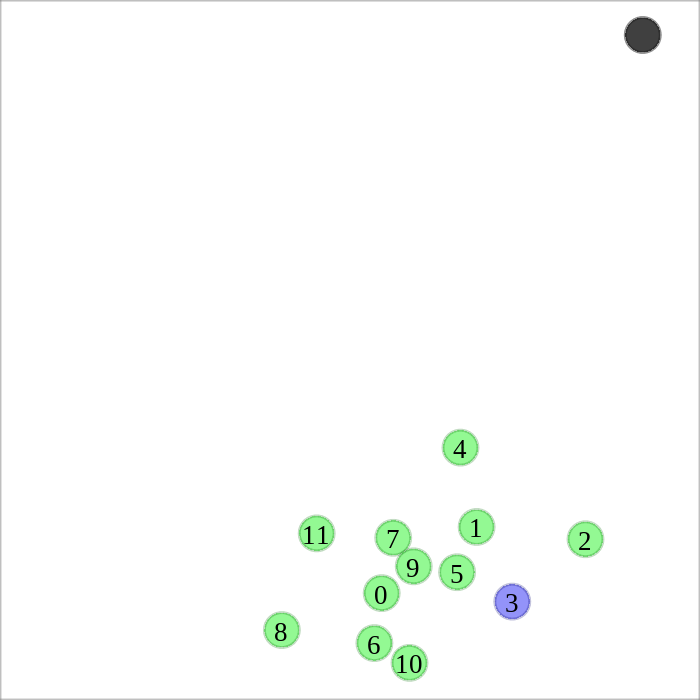}} 
         \fbox{\includegraphics[width=\www]{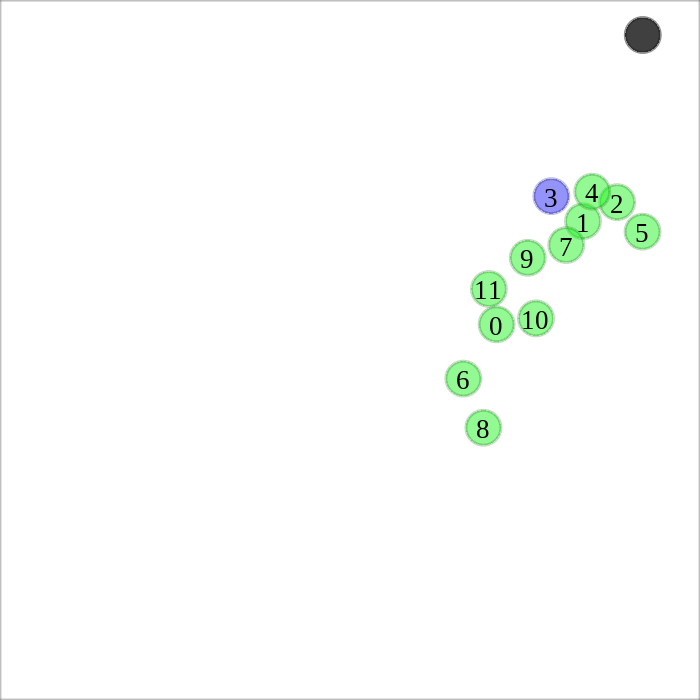}}
         \fbox{\includegraphics[width=\www]{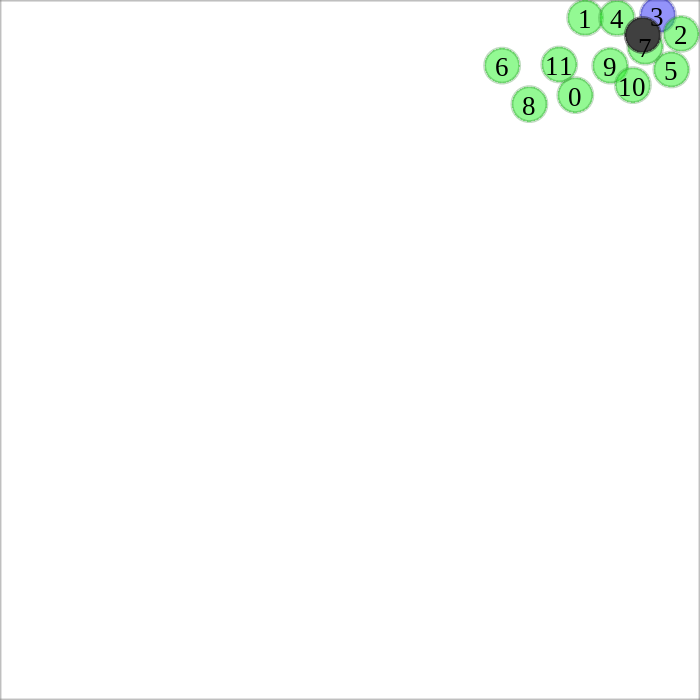}}
    \caption{\small 0-shot generalization of our proposed approach to different multi-agent team size. We trained a multi-agent policy with $n=6$ agents using our approach directly tested the performance with $2n=12$ agents without any fine-tuning. The multi-agent team with double the number of agents in able to navigate to the goal location while successfully hiding the leader within the followers.}
    \label{fig:large_swarm}
\end{figure*}

\begin{figure*}[ht]
     \centering
     \begin{subfigure}[b]{0.3\textwidth}
         \centering
         \includegraphics[width=\textwidth]{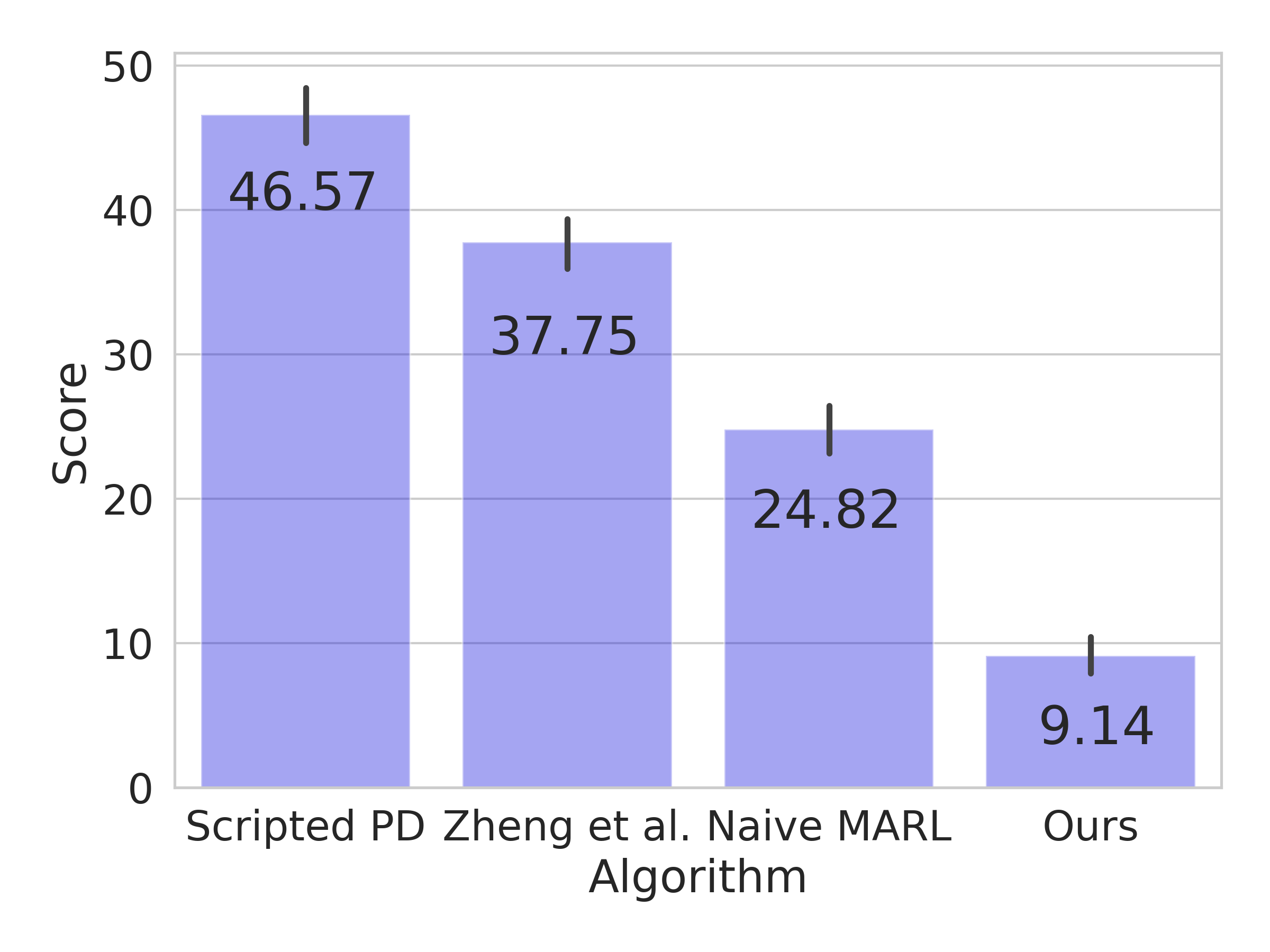}
         \vspace{-7mm}
         \caption{Humans' Score}
         \label{fig:score}
     \end{subfigure}
     \begin{subfigure}[b]{0.3\textwidth}
         \centering
         \includegraphics[width=\textwidth]{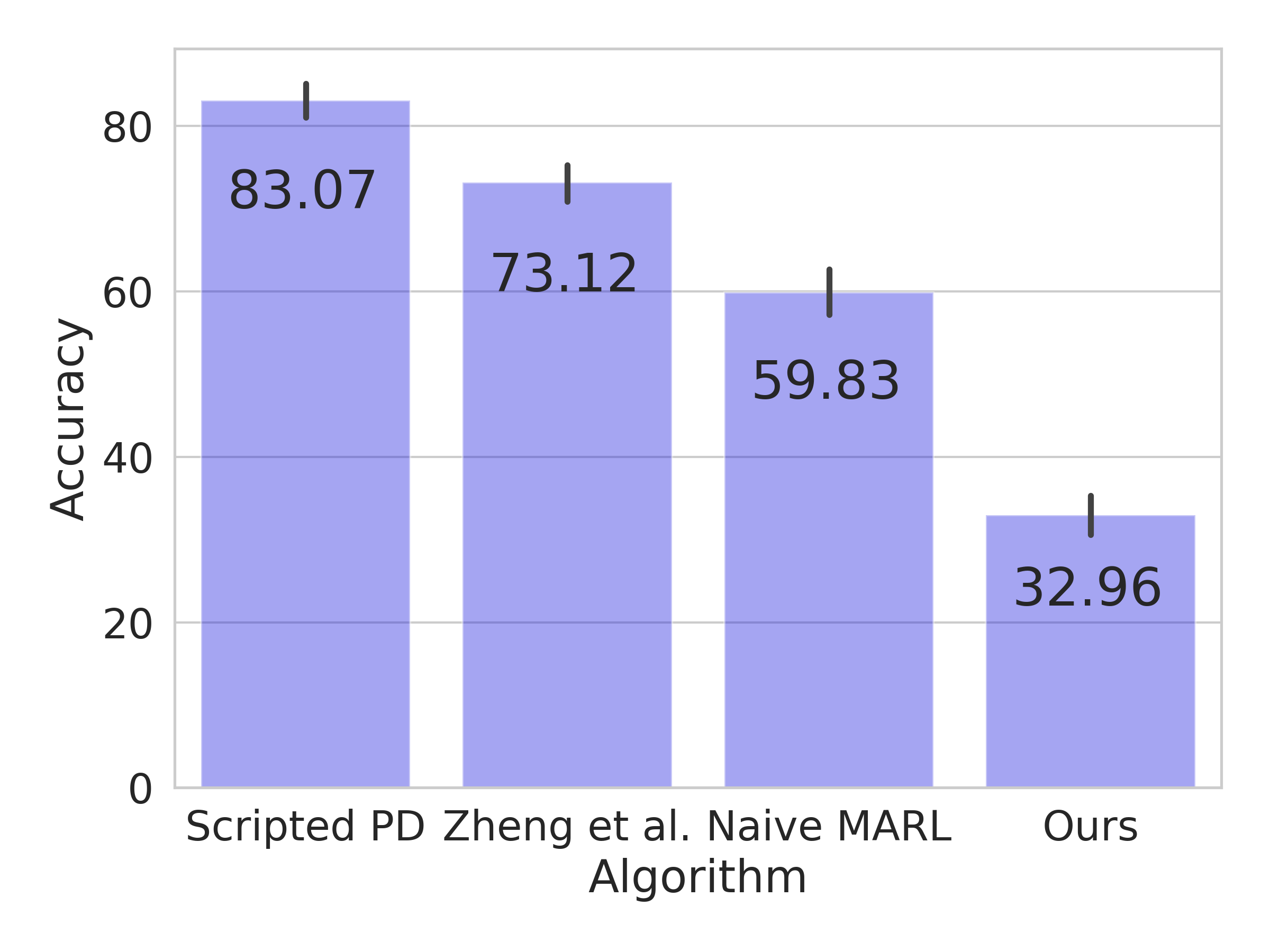}
         \vspace{-7mm}
         \caption{Humans' Accuracy}
         \label{fig:accuracy}
     \end{subfigure}
     \begin{subfigure}[b]{0.3\textwidth}
         \centering
         \includegraphics[width=\textwidth]{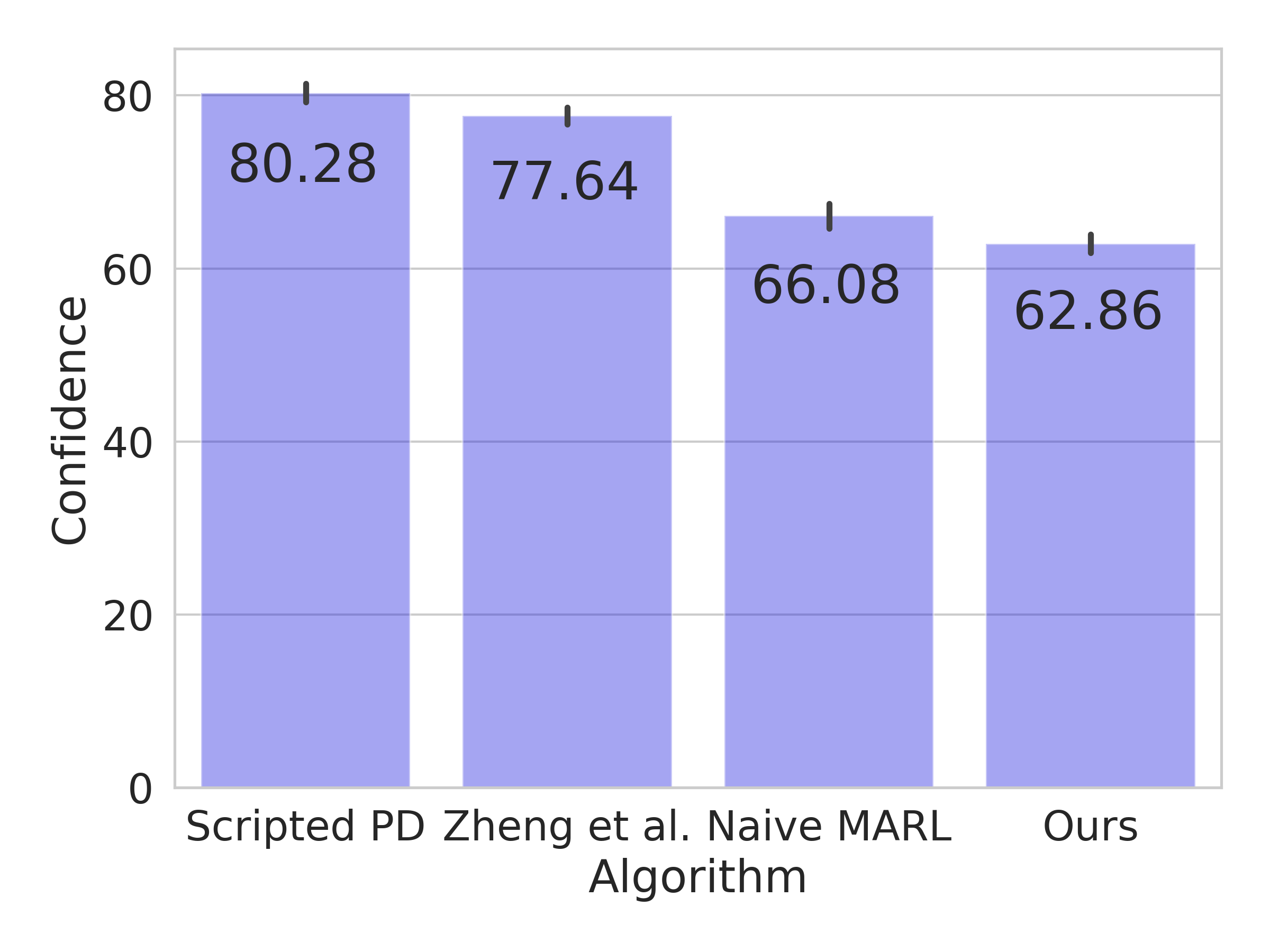}
         \vspace{-7mm}
         \caption{Humans' Confidence}
         \label{fig:confidence}
     \end{subfigure}
    \caption{\small 
    This figure presents comparative results of human observers (adversary) trying to identify the leader in multi-robot teams executing different navigation algorithms (Scirpted PD, Zheng et al. \cite{zheng2020adversarial}, Naive MARL and ours). The results show that our algorithm performs better (\textbf{lower value is better}) in terms of overall score (\ref{fig:score}), accuracy (\ref{fig:accuracy}) and confidence (\ref{fig:confidence}).}
    \label{fig:human_data}
    \vspace{-6mm}
\end{figure*}

We first show that our proposed Scalable-LSTM adversary architecture performs extremely well in identifying the leader when the multi-robot team doesn't try to hide it's leader's identity. This is important because there would be no point in designing and evaluating a multi-agent policy against a weak adversary.

We collect 100 episodes of multi-agent (6 agents) trajectories from a Naive MARL policy (Section \ref{section:training_state_1}) and test the performance of Scalable-LSTM along with other baselines in identifying the leader correctly. The baselines includes random guess and deep learning adversary architectures LSTM, \cite{hochreiter1997long} and Zheng et al. \cite{zheng2020adversarial} which has a Convolutional Neural Network (CNN) architecture. Scalable-LSTM has an accuracy of 0.99 which is higher than that of LSTM (0.95) and Zheng et al. (0.97) as shown in Table \ref{tab:adversary_comparision}. Table \ref{tab:adversary_comparision} also shows that Scalable-LSTM has 2 orders of magnitude lesser parameters than Zheng et al. \cite{zheng2020adversarial} and still performs better.

Another advantage of Scalabe-LSTM over the baseline deep learning architectures is that it is adaptable both temporally and in the number of agents, which is appealing for real world applications as an adversary might observe only a part of the trajectory or the multi-agent team size might change. 
\vspace{2mm}
\subsubsection{Temporal adaptability}
We show an example episode in Fig. \ref{fig:adversary_temporal_adaptability} to demonstrate temporal adaptability. The adversary can make a prediction of the leader's identity at every time step, i.e. even from partial trajectories. Within 10 time steps, the adversary correctly identifies the leader and as time flows it predicts the true leader with higher confidence (assigns higher probability to true leader).

\subsubsection{Adaptability in no. of agents}
We trained the Scalable-LSTM adversary to identify the leader in 6-agent team and tested it on 3 to 10 agent teams without any fine tuning on different team sizes. Leader prediction accuracy remains high in all cases as shown in Fig. \ref{fig:adversary_num_agents_generalization}. Thus Scalable-LSTM shows 0-shot generalization to different number of agents.


\subsection{Multi-Agent Performance}
\label{section:results_swarm_performance}
\subsubsection{Baselines}
We test the multi-agent policy's performance based on primary task reward (efficiently reaching the goal) and and on its ability to hide the leader's identity from artificial adversary (identity hiding reward). These rewards are only used for evaluation purpose and are not accessible to the agents at test time.
We compare the performance against traditional leader-follower controllers - Scripted PD and Zheng et al. \cite{zheng2020adversarial} as well as Naive MARL policy (Section \ref{section:training_state_1}). Scripted PD is a Proportional-Differential (PD) controller based algorithm where the leader tracks the goal location and the followers track the leader. Zheng et al. \cite{zheng2020adversarial} used traditional control based method to track waypoints in the trajectory. In their method, the followers don't know the leader's identity. We adopted their method to track the single goal instead of multiple waypoints. Scripted PD and Naive MARL only try to reach the goal location while Zheng et al. \cite{zheng2020adversarial} and our approaches also try to hide the leader's identity.

\subsubsection{Goal reaching task}
On the goal reaching task, Fig. \ref{fig:task_reward}, all the methods we tested performed well and could successfully navigate the swarm to its goal. The slight drop in performance of our approach and Zheng et al. \cite{zheng2020adversarial} is expected as these methods might have to trade-off primary task objective with identity hiding objective. Among the methods that considered identity hiding objective, our approach got higher reward than Zheng et al. \cite{zheng2020adversarial}.

\subsubsection{Leader Identity hiding}
In terms of identity hiding reward, Fig. \ref{fig:privacy_reward}, our approach significantly outperformed all the baselines, hiding the leader successfully in most of the cases. 

Fig. \ref{fig:swarm_trajectories} shows example multi-robot trajectories using the approach of Zheng et al. \cite{zheng2020adversarial} and our approach. In case of Zheng et al. \cite{zheng2020adversarial}, the leader clearly leads the followers and this makes it easy for an adversary to identify the leader, Fig. \ref{fig:zheng_trajectories}. Using our approach, the leader doesn't naively lead the other agents. It rather smartly moves together with the group while informing the target motion direction through hidden movement cues. This smart strategy helps our multi-agent team deceive the adversary while achieving the primary goal, Fig. \ref{fig:our_trajectories}. We show successful leader-follower navigation to different goal locations while hiding the leader's identity using our proposed approach \ref{fig:different_goals}. 

\subsubsection{0-shot generalization}
\label{section:swarm_generalization_num_agents}
We demonstrate the 0-shot generalization of our multi-agent policy to different number of agents in Fig. \ref{fig:large_swarm}, where we train a policy on $n=6$ agents and test on $2n=12$ agents without any fine-tuning.

\subsection{Effectiveness against Human observers}
Although we have shown that our trained multi-agent policy can successfully hide its leader's identity from artificial adversary, we wish to push it further and test how well the learnt navigation strategy can deceive humans when they act as adversaries and try to identify the leader. In the following sections we describe the experimental setup and the results.
\subsubsection{Experimental setup}
We tasked humans with identifying the leader in a multi-agent team by showing videos of leader-follower multi-robot navigation with visually identical leader and followers.
The human's goal was to maximize a score that incentivized answers that were both correct and quick. For each algorithm (our and baselines), we collected data from $\sim30$ humans and showed $50$ videos to each human. In total, we collected $\sim 30 \times 50 = 1500$ data points for each algorithm to get a good estimate of its performance. We collected our data through a web based interface
via Amazon Mechanical Turk (MTurk) \cite{turk2012amazon}. Amazon Mechanical Turk is a crowd-sourcing platform where real humans are paid money to perform tasks.

\subsubsection{Results}
Fig. \ref{fig:human_data} shows the performance of different algorithms - Scripted PD, Zheng et al. \cite{zheng2020adversarial}, Naive MARL and ours against human adversaries. From a multi-robot team's perspective, a lower score achieved by humans is indicative of better performance of multi-robot team. Across all the metrics, our algorithm outperformed all of the other algorithms. Humans got much lower score (Fig. \ref{fig:score}), lower accuracy (Fig. \ref{fig:accuracy}) and lacked confidence (Fig. \ref{fig:confidence}) in their decisions. It is interesting to note that the Zheng et al. \cite{zheng2020adversarial} didn't generalize to human data well and performed worse than the Naive MARL policy we trained. On the other hand, our algorithm generalized well to completely unseen humans making it hard for humans to infer the leader's identity.


\section{Conclusion and future work}
In this paper we brought together MARL, GNNs and adversarial training for the task of hiding the leader's identity in a multi-robot team as a defense mechanism.
Our proposed leader-follower navigation algorithm allows decentralized control, adapts to a variable number of agents and generalizes to unseen human adversaries showing the effectiveness of our algorithm.
Given the formulation of leader identity hiding as a MARL problem, other leader-follower tasks (such as trajectory following) could also be adopted by changing the task-related reward structure of MARL.

\addtolength{\textheight}{-12cm}   








\bibliographystyle{IEEEtran}
\bibliography{references.bib}

\begin{thebibliography}{10}
\providecommand{\url}[1]{#1}
\csname url@samestyle\endcsname
\providecommand{\newblock}{\relax}
\providecommand{\bibinfo}[2]{#2}
\providecommand{\BIBentrySTDinterwordspacing}{\spaceskip=0pt\relax}
\providecommand{\BIBentryALTinterwordstretchfactor}{4}
\providecommand{\BIBentryALTinterwordspacing}{\spaceskip=\fontdimen2\font plus
\BIBentryALTinterwordstretchfactor\fontdimen3\font minus
  \fontdimen4\font\relax}
\providecommand{\BIBforeignlanguage}[2]{{%
\expandafter\ifx\csname l@#1\endcsname\relax
\typeout{** WARNING: IEEEtran.bst: No hyphenation pattern has been}%
\typeout{** loaded for the language `#1'. Using the pattern for}%
\typeout{** the default language instead.}%
\else
\language=\csname l@#1\endcsname
\fi
#2}}
\providecommand{\BIBdecl}{\relax}
\BIBdecl

\bibitem{choudhury2021efficient}
S.~Choudhury, K.~Solovey, M.~J. Kochenderfer, and M.~Pavone, ``Efficient
  large-scale multi-drone delivery using transit networks,'' \emph{Journal of
  Artificial Intelligence Research}, vol.~70, pp. 757--788, 2021.

\bibitem{albani2017monitoring}
D.~Albani, J.~IJsselmuiden, R.~Haken, and V.~Trianni, ``Monitoring and mapping
  with robot swarms for agricultural applications,'' in \emph{2017 14th IEEE
  International Conference on Advanced Video and Signal Based Surveillance
  (AVSS)}.\hskip 1em plus 0.5em minus 0.4em\relax IEEE, 2017, pp. 1--6.

\bibitem{queralta2020collaborative}
J.~P. Queralta, J.~Taipalmaa, B.~C. Pullinen, V.~K. Sarker, T.~N. Gia,
  H.~Tenhunen, M.~Gabbouj, J.~Raitoharju, and T.~Westerlund, ``Collaborative
  multi-robot search and rescue: Planning, coordination, perception, and active
  vision,'' \emph{IEEE Access}, vol.~8, pp. 191\,617--191\,643, 2020.

\bibitem{gregory2016application}
J.~Gregory, J.~Fink, E.~Stump, J.~Twigg, J.~Rogers, D.~Baran, N.~Fung, and
  S.~Young, ``Application of multi-robot systems to disaster-relief scenarios
  with limited communication,'' in \emph{Field and Service Robotics}.\hskip 1em
  plus 0.5em minus 0.4em\relax Springer, 2016, pp. 639--653.

\bibitem{deka2021natural}
A.~Deka and K.~Sycara, ``Natural emergence of heterogeneous strategies in
  artificially intelligent competitive teams,'' in \emph{International
  Conference on Swarm Intelligence}.\hskip 1em plus 0.5em minus 0.4em\relax
  Springer, 2021, pp. 13--25.

\bibitem{zheng2020adversarial}
H.~Zheng, J.~Panerati, G.~Beltrame, and A.~Prorok, ``An adversarial approach to
  private flocking in mobile robot teams,'' \emph{IEEE Robotics and Automation
  Letters}, vol.~5, no.~2, pp. 1009--1016, 2020.

\bibitem{sakai2017leader}
D.~Sakai, H.~Fukushima, and F.~Matsuno, ``Leader--follower navigation in
  obstacle environments while preserving connectivity without data
  transmission,'' \emph{IEEE Transactions on Control Systems Technology},
  vol.~26, no.~4, pp. 1233--1248, 2017.

\bibitem{vilca2016adaptive}
J.~Vilca, L.~Adouane, and Y.~Mezouar, ``Adaptive leader-follower formation in
  cluttered environment using dynamic target reconfiguration,'' in
  \emph{Distributed Autonomous Robotic Systems}.\hskip 1em plus 0.5em minus
  0.4em\relax Springer, 2016, pp. 237--254.

\bibitem{simonsenapplication}
A.~S. Simonsen and E.-L.~M. Ruud, ``The application of a flexible
  leader-follower control algorithm to different mobile autonomous robots.''

\bibitem{miah2020model}
M.~S. Miah, A.~Elhussein, F.~Keshtkar, and M.~Abouheaf, ``Model-free
  reinforcement learning approach for leader-follower formation using
  nonholonomic mobile robots,'' in \emph{The Thirty-Third International Flairs
  Conference}, 2020.

\bibitem{zhou2019adaptive}
Y.~Zhou, F.~Lu, G.~Pu, X.~Ma, R.~Sun, H.-Y. Chen, and X.~Li, ``Adaptive
  leader-follower formation control and obstacle avoidance via deep
  reinforcement learning,'' in \emph{2019 IEEE/RSJ International Conference on
  Intelligent Robots and Systems (IROS)}.\hskip 1em plus 0.5em minus
  0.4em\relax IEEE, 2019, pp. 4273--4280.

\bibitem{amato2013decentralized}
C.~Amato, G.~Chowdhary, A.~Geramifard, N.~K. {\"U}re, and M.~J. Kochenderfer,
  ``Decentralized control of partially observable markov decision processes,''
  in \emph{52nd IEEE Conference on Decision and Control}.\hskip 1em plus 0.5em
  minus 0.4em\relax IEEE, 2013, pp. 2398--2405.

\bibitem{deka2018adaptive}
A.~Deka, V.~K. Narayanan, T.~Miyashita, and N.~Hagita, ``Adaptive
  attention-aware pedestrian trajectory prediction for robot planning in human
  environments,'' \emph{Unpublished}, 2018.

\bibitem{vaswani2017attention}
A.~Vaswani, N.~Shazeer, N.~Parmar, J.~Uszkoreit, L.~Jones, A.~N. Gomez,
  L.~Kaiser, and I.~Polosukhin, ``Attention is all you need,'' \emph{arXiv
  preprint arXiv:1706.03762}, 2017.

\bibitem{hochreiter1997long}
S.~Hochreiter and J.~Schmidhuber, ``Long short-term memory,'' \emph{Neural
  computation}, vol.~9, no.~8, pp. 1735--1780, 1997.

\bibitem{schulman2017proximal}
J.~Schulman, F.~Wolski, P.~Dhariwal, A.~Radford, and O.~Klimov, ``Proximal
  policy optimization algorithms,'' \emph{arXiv preprint arXiv:1707.06347},
  2017.

\bibitem{lowe2017multi}
R.~Lowe, Y.~Wu, A.~Tamar, J.~Harb, P.~Abbeel, and I.~Mordatch, ``Multi-agent
  actor-critic for mixed cooperative-competitive environments,'' \emph{arXiv
  preprint arXiv:1706.02275}, 2017.

\bibitem{turk2012amazon}
A.~M. Turk, ``Amazon mechanical turk,'' \emph{Retrieved August}, vol.~17, p.
  2012, 2012.

\end{thebibliography}
\end{document}